\documentclass{trbunofficial}
\usepackage{graphicx}
\usepackage{tabularx}
\usepackage{xcolor}
\usepackage{soul}
\usepackage{appendix}

\usepackage[hidelinks]{hyperref}

\usepackage{booktabs}
\usepackage{subcaption}
\usepackage{multirow}

\usepackage{amsfonts}
\usepackage{color} 
\usepackage{bbding}

\AuthorHeaders{Guo, Zhang, et al.}

\title{Towards Explainable Traffic Flow Prediction with Large Language Models}

\author{%
  \textbf{Xusen Guo, Co-first authors}\\
     Ph.D. Student\\
     Systems Hub, Intelligent Transportation Thrust\\
     Hong Kong University of Science and Technology (Guangzhou), Guangzhou, China\\
     xguo796@connect.hkust-gz.edu.cn\\
    \hfill\break%
\textbf{Qiming Zhang, Co-first authors}\\
     Ph.D. Student\\
     Systems Hub, Intelligent Transportation Thrust\\
     Hong Kong University of Science and Technology (Guangzhou), Guangzhou, China\\
     qzhang255@connect.hkust-gz.edu.cn\\
    \hfill\break%
\textbf{Junyue Jiang}\\
    Ph.D. Student\\
    Department of Civil and System Engineering\\
    Johns Hopkins University\\
    jjiang67@jhu.edu\\
    \hfill\break%
\textbf{Mingxing Peng}\\
    Ph.D. Student\\
    Systems Hub, Intelligent Transportation Thrust\\
    Hong Kong University of Science and Technology (Guangzhou), Guangzhou, China\\
    mpeng060@connect.hkust-gz.edu.cn\\
    \hfill\break
\textbf{Meixin Zhu, Ph.D., Corresponding Author}\\
    Assistant Professor\\
    Systems Hub, Intelligent Transportation Thrust\\
    Hong Kong University of Science and Technology (Guangzhou), Guangzhou, China\\
    meixin@ust.hk\\
    \hfill\break%
\textbf{Hao (Frank) Yang, Ph.D., Corresponding Author}\\
    Assistant Professor\\
    Department of Civil and System Engineering\\
    Johns Hopkins University\\
    haofrankyang@jhu.edu
    \hfill\break%
}

\begin{document}
\maketitle
\section{Abstract}
Traffic forecasting is crucial for intelligent transportation systems. It has experienced significant advancements thanks to the power of deep learning in capturing latent patterns of traffic data. However, recent deep-learning architectures require intricate model designs and lack an intuitive understanding of the mapping from input data to predicted results. Achieving both accuracy and explainability in traffic prediction models remains a challenge due to the complexity of traffic data and the inherent opacity of deep learning models. To tackle these challenges, we propose a \textbf{T}raffic flow \textbf{P}rediction model based on \textbf{L}arge \textbf{L}anguage \textbf{M}odels (LLMs) to generate e\textbf{x}plainable traffic predictions, named \textbf{xTP-LLM}. By transferring multi-modal traffic data into natural language descriptions, xTP-LLM captures complex time-series patterns and external factors from comprehensive traffic data. The LLM framework is fine-tuned using language-based instructions to align with spatial-temporal traffic flow data. Empirically, xTP-LLM shows competitive accuracy compared with deep learning baselines, while providing an intuitive and reliable explanation for predictions. This paper contributes to advancing explainable traffic prediction models and lays a foundation for future exploration of LLM applications in transportation. To the best of our knowledge, this is the first study to use LLM for explainable prediction of traffic flows.

\hfill\break%
\noindent\textit{Keywords}: Traffic flow prediction, Large language models, Spatial-temporal prediction, Explainability.
\newpage

\section{Introduction}\label{I}
Traffic network prediction is a critical component of transportation management systems, aiming to forecast future traffic conditions such as congestion, traffic volume, and travel time. It plays a vital role in various applications, including route planning, traffic management, and intelligent transportation systems (ITS) \cite{dimitrakopoulos2010intelligent}. Accurate predictions are crucial for providing valuable insights to stakeholders in traffic systems, aiding in informed decision-making. However, achieving reliability and precision in predictions is challenging due to the inherent nonlinear dynamics of traffic, spatial and temporal variability, and dynamic nature influenced by factors like accidents, weather conditions, and other events, necessitating models capable of capturing complex dependencies and rapid changes. As large-scale traffic data becomes more available and deep learning techniques advance, data-driven methods are increasingly favored for modeling complex traffic flow systems.

Deep learning-based traffic analysis \cite{wang2020deep} have been well researched, covering human mobility research \cite{zhang2017deep, Jin_Lin_Wu_Wan_2018}, traffic management \cite{du2018hybrid, ranjan2020city}, and accident analysis \cite{yannis2017road}. Typically, these issues are treated as spatio-temporal deep learning problems. Deep learning methods consistently learn hierarchical feature representations from spatial-temporal data, understand historical trends \cite{Yang_Dillon_Chen_2017, fan2018online}, and employ graphs to illustrate the spatial relationships between locations \cite{marblestone2016toward, gao2018trajectory}. The spatio-temporal-graph learning paradigm is the primary methodology for learning representations and capturing potential trends and relationships from traffic data.

Despite the comprehensive model architectures, this domain format still faces challenges. First, deep learning methods require specific network structure designs \cite{Li_Yu_Shahabi_Liu_2017, akbari2018short}, to consider the multi-modal dynamic nature and spatio-temporal complexity of traffic data. Although these designs can help the model improve the accuracy of predictions, they are generally artificially crafted, and abstract representations further obscure generalization ability. Additionally, some traffic forecasting studies \cite{dai2020nonlinear, zhou2019reliable, chen2020acting} related to the reliability and robustness of models prioritize the degree of fit to actual values. They consider a variety of dynamic factors affecting traffic flow to improve the reliability of traffic flow predictions. However, these methodologies encounter difficulties in providing credentials for prediction and taking accountability for outcomes in real-world traffic prediction tasks that lack labeled data.

\begin{figure}[!h]
  \centering\includegraphics[width=0.8\columnwidth]{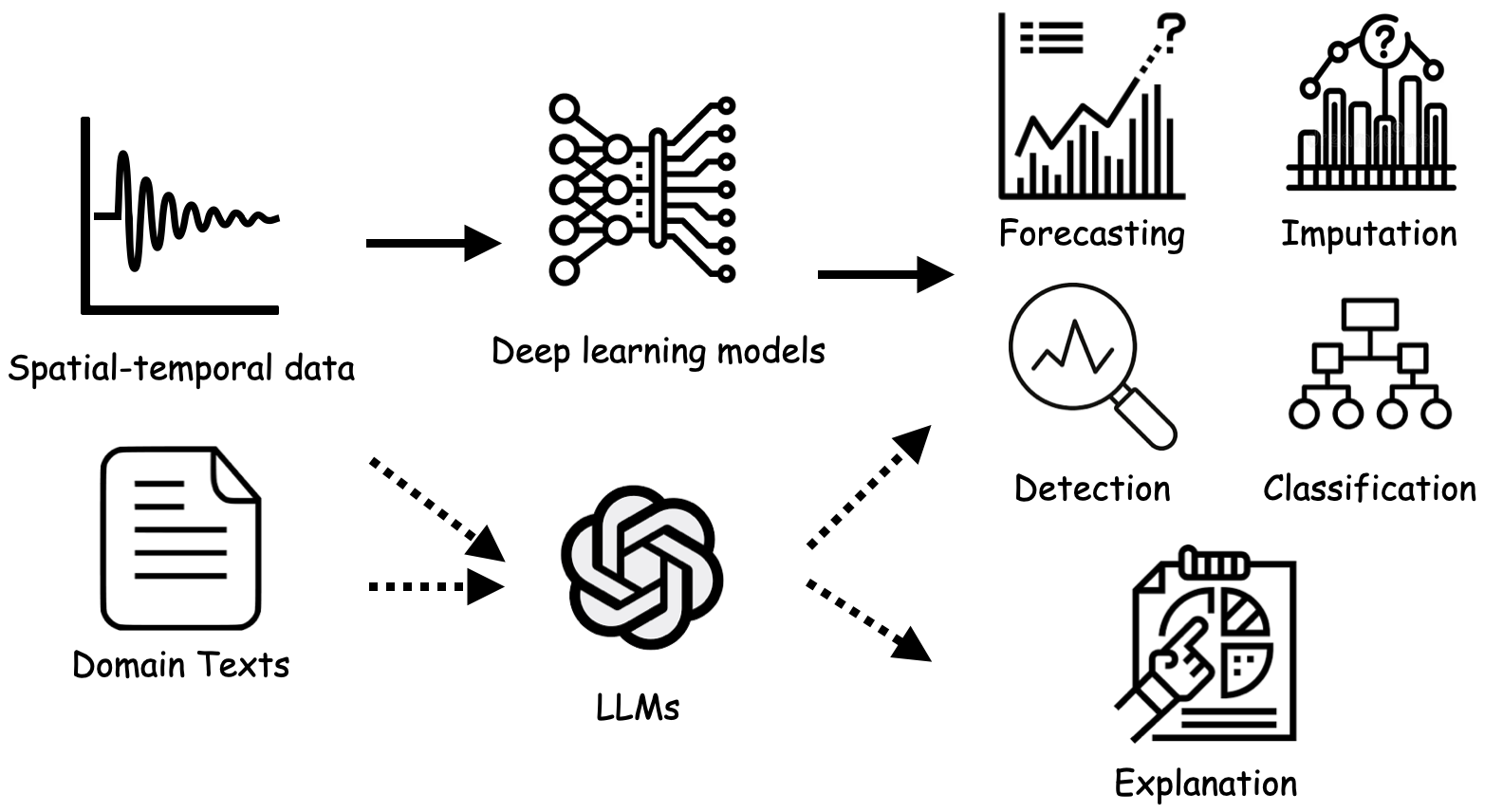}
  \caption{Spatial-temporal learning formats for deep learning models and LLMs: Compared with well-developed deep learning models, LLMs for spatial-temporal learning advance in adapting domain knowledge with urban multi-modal data and generate reasonable explanations. }
  \label{fig:fig0}
\end{figure}

Recently, with the popularity of foundation models \cite{chang2023examining, touvron2023llama}, large language models provide more intuitive explainable trials in spatial-temporal domain tasks, benefiting from textual paradigms. Spatial-temporal learning tasks can be refined into language format for exploring the potential of LLMs in various application fields, including forecasting \cite{jin2023time, shi2024language}, data imputation \cite{chen2023gatgpt}, and anomaly detection \cite{wan2024dell}, as depicted in Figure~\ref{fig:fig0}. Considering multi-modalities in urban big data, the LLM framework transfers original data into natural language description and is capable of capturing latent relationships between inputs from complicated contexts. Also, LLMs can generate explanations of the reasoning process, which provides available supplements for prediction and decision-making. However, though LLMs have expertise in language-based tasks, how to formulate specific spatial-temporal tasks into LLM frameworks for accurate prediction remains challenging.

In response to the challenges mentioned, we present xTP-LLM, a novel traffic flow prediction framework utilizing large language models. This framework effectively predicts future traffic flow using multi-modal data while providing interpretable explanations. By converting traffic data and external factors—such as points of interest, weather conditions, dates, and holidays—into a structured prompt, xTP-LLM captures domain-specific information more effectively. Our experiments reveal that xTP-LLM not only outperforms state-of-the-art deep learning models in accuracy but also exhibits robust generalization across various scenarios. The main contributions of this work are as follows:
\begin{itemize}

    \item We design a structured textual prompt that incorporates multi-modal traffic flow information, facilitating LLMs to capture traffic patterns better. Additionally, we reorganized the multi-modal traffic flow data in California to create a comprehensive text-based dataset, CATraffic, for future exploration in traffic prediction with LLMs. 
    
    \item We propose xTP-LLM, a traffic flow prediction framework based on large language models. This framework exhibits competitive accuracy against state-of-the-art deep learning models and demonstrates effective generalization abilities across different traffic flow prediction scenarios without additional training.
    
    \item xTP-LLM can provide insightful explanations for its prediction results, offering greater interpretability than traditional deep learning approaches, which facilitates rational decision-making in traffic management and planning.
    
\end{itemize}

For the rest parts of this paper, the related works are involved in Section II, and details about methodology are illustrated in Section III. Comparison results, ablation, explanation, and what-if studies are included in experimental results and analysis, as Section IV. Section V summarizes this paper and gives insights for future exploration of LLM applications in transportation.

\section{Related Work}

In this section, we will first explore advancements in spatial-temporal prediction, emphasizing the integration of deep learning methodologies. Subsequently, we'll delve into the importance of reliable prediction, discussing methods for enhancing explainability in spatial-temporal learning. Finally, we'll highlight the transformative role of LLMs across diverse domains, elucidating their pre-training and fine-tuning practices for domain-specific tasks.

\subsection{Spatial-temporal prediction}
Recently, the field of spatial-temporal learning has witnessed significant advancements, especially in the traffic, environment, and society field, due to the emergence of deep learning methodologies. These approaches have enabled the modeling of latent relationships among various features of urban data in diverse formats. These architectural designs are meticulously crafted to comprehend and represent the intricate interplay between spatial and temporal dimensions within datasets. Convolutional Neural Networks(CNNs) \cite{zhang2017deep, liang2018geoman}, quite renowned for their efficacy in computer vision, are employed to discern the spatial relations among grid regions, by filtering the input data. Moreover, Recurrent Neural Networks (RNNs) \cite{graves2012long} are usually leveraged to adeptly capture temporal dependencies, through the maintenance of a memory state, facilitating the reusing of information over time. Notably, more spatial-temporal learning frameworks introduce Graph Neural Networks(GNNs) \cite{wu2019graph, yu2017spatio, guo2019attention}, advanced in the representation of complex spatial relationships inherent in data structured as graphs, wherein nodes correspond to spatial locations and edges encapsulate the connections between them. Additionally, the adaptation of Transformers \cite{vaswani2017attention}, originally proposed for natural language processing, has proven effective in long-sequence modeling to capture comprehensive information. Various Transformer blocks \cite{yu2020spatio, cai2020traffic} have been tailored for different dependencies among spatial-temporal features, enabling the modeling of intricate relationships. A notable trend in this domain is the combination of different model architectures, leveraging various modules for spatial or temporal features \cite{xu2020spatial, jiang2023pdformer}. This amalgamation gradually becomes the prevailing paradigm, showcasing promising performance in prediction tasks. However, it's worth noting that while these methods perform excellent in prediction accuracy, they often fall short in terms of explainability and generalization.

\subsection{Explainable prediction}
The explainability of spatial-temporal learning is also worthy of consideration for reliable prediction, which provides abundant views beyond prediction accuracy. Most recent works studied which features mostly affect decisions generated by models. \cite{barredo2019lies} focuses on the dependency on latent variables of road forecasting based on black-box machine learning methods, including RNNs and Random Forests. The spatial-temporal causal graph inference, as presented in reference \cite{zhang2022granger}, offers an approximation of the Granger causality test, thereby enhancing the accessibility of forecasting. Counterfactual explanations for time series \cite{yan2023self, ates2021counterfactual} are also highly regarded, as they concentrate on generating alternative prediction outcomes by selecting time series data points from the training set and substituting them into the sample under analysis. This method allows for the illustration of results by examining a limited number of variables. Besides these methods focusing on model transparency, large language models offer an alternative approach to generate convincible explanations along with prediction results with greater intuitiveness \cite{huang2023can, peng2024lc, gruver2024large}. As a bridge between complex systems and humans, LLMs have the potential to convert input-output mappings into natural, easy-to-understand narratives, helping humans make more informed and reliable decisions\cite{zytek2024llms, cambria2024xai}.

\subsection{Large language models}
Large Language Models (LLMs) have achieved remarkable success across a wide range of tasks and fields, including natural language processing \cite{ray2023chatgpt}, vision-language tasks \cite{liu2023llava}, and various other interdisciplinary domains \cite{thirunavukarasu2023large, da2024prompt, wu2023bloomberggpt}. Originally designed as pre-trained language foundation models for addressing various natural language tasks, LLMs have exhibited the capacity to acquire intricate semantic and knowledge representations from extensive text corpora over time. This newfound ability has been a profound source of inspiration within the community for addressing a variety of tasks. The success of models like GPT-4 \cite{chang2023examining} in natural language understanding and generation tasks has spurred interest in exploring their potential for handling complex, multi-modal datasets beyond traditional linguistic domains. They can extract valuable information and relationships from complex textual contexts, thereby enhancing the learning of city data. With the popularity of decoder-architecture LLMs, domain tasks are normally formulated into the next token generation, which provides a unified formulation to learn the map from the input to the output. To acquire large models for specific fields, the practices of pre-training and fine-tuning have become widely accepted in the model training process. Pre-training a foundation model from scratch necessitates substantial computing resources and domain-specific datasets, resulting in its superior performance within professional domains compared to baseline models. On the other hand, fine-tuning based on foundation models offers a more accessible approach, involving adjustments to only a few parameters \cite{hu2022lora}. This method preserves most general knowledge while targeting expertise in domain-specific tasks. In some cases, researchers freeze all parameters of large language models and focus solely on training the extended encoders and decoders \cite{jin2023time, zhu2023minigpt}. This strategy aims to extend the learning capabilities of LLMs to domain-specific tasks while leveraging their existing knowledge base.

\section{Methodology}

This paper presents a novel approach utilizing Large Language Models for traffic flow prediction. Our objective is to develop a predictive model that not only forecasts traffic flow patterns but also provides explanations for the predicted trends. In the following sections, we will provide a comprehensive overview of our approach. Firstly, we'll describe the problem formulation and the predictive framework. Next, we'll discuss the construction of prompts, which is crucial for fine-tuning Large Language Models. Finally, we'll talk about the fine-tuning technique and how to generate the explanation corresponding to the predicted result.

\subsection{Problem Description}
The traffic flow prediction problem as a part of time-series prediction problems, can be formulated as forecasting future values according to the historical data. In our framework, the goal is to predict future-step values and generate explanations based on historical values and external factors. We can represent this using the following formula:
\begin{align}
\quad X_{T:T+H}, \mathbb{I}_i = P_{\theta}(X_{T-H-1:T-1}, E_i)
\end{align}
Here, the function $P_{\theta}$ is the predictive model, that learns the mapping relationship between input and output. $X_{T-H-1:T-1}$, $X_{T:T+H}$ represent continuous historical values and future predicted values with $H$ steps. $E_i$ is external factors for data sample $i$, which includes date, holiday information,  meteorological data and PoI data, etc. $\mathbb{I}_i$ is the explanation corresponding to the predicted output $X_{T:T+H}$. In this work, we treat the large language model as the predictive model $P_{\theta}$, and utilize the language tokenizer to transform the input and output data as a sequence of tokens, thus reframing the task of traffic flow prediction as a language modeling problem. Specifically, given the tokenized input sequence $S_{T-H-1:T-1}$ and tokenized external factors $SE_i$ for data sample $i$, the tokenized output sequence $S_{T:T+H}$ is reconstructed autoregressively:
\begin{align}
\hat{s}_{T+i}  = \mathop{\arg\max}\limits_{s_{T+i}}  P_{\theta}(s_{T+i}|S_{T:T+i-1}, S_{T-H-1:T-1}, SE_i)
\end{align}
Here, $\hat{s}_{T+i}$ represents the $i$-th predicted next token in output sequence. Similarly, the explanation can also be generated following the output sequence in this way. By incorporating the next token prediction into the framework, we enhance the language model's ability to generate coherent and contextually relevant explanations alongside traffic flow predictions. 

\subsection{Prompt Construction}

\begin{figure}[!h]
        \centering
        \includegraphics[trim={2cm 5cm 0cm 3cm}, clip,width=\linewidth]{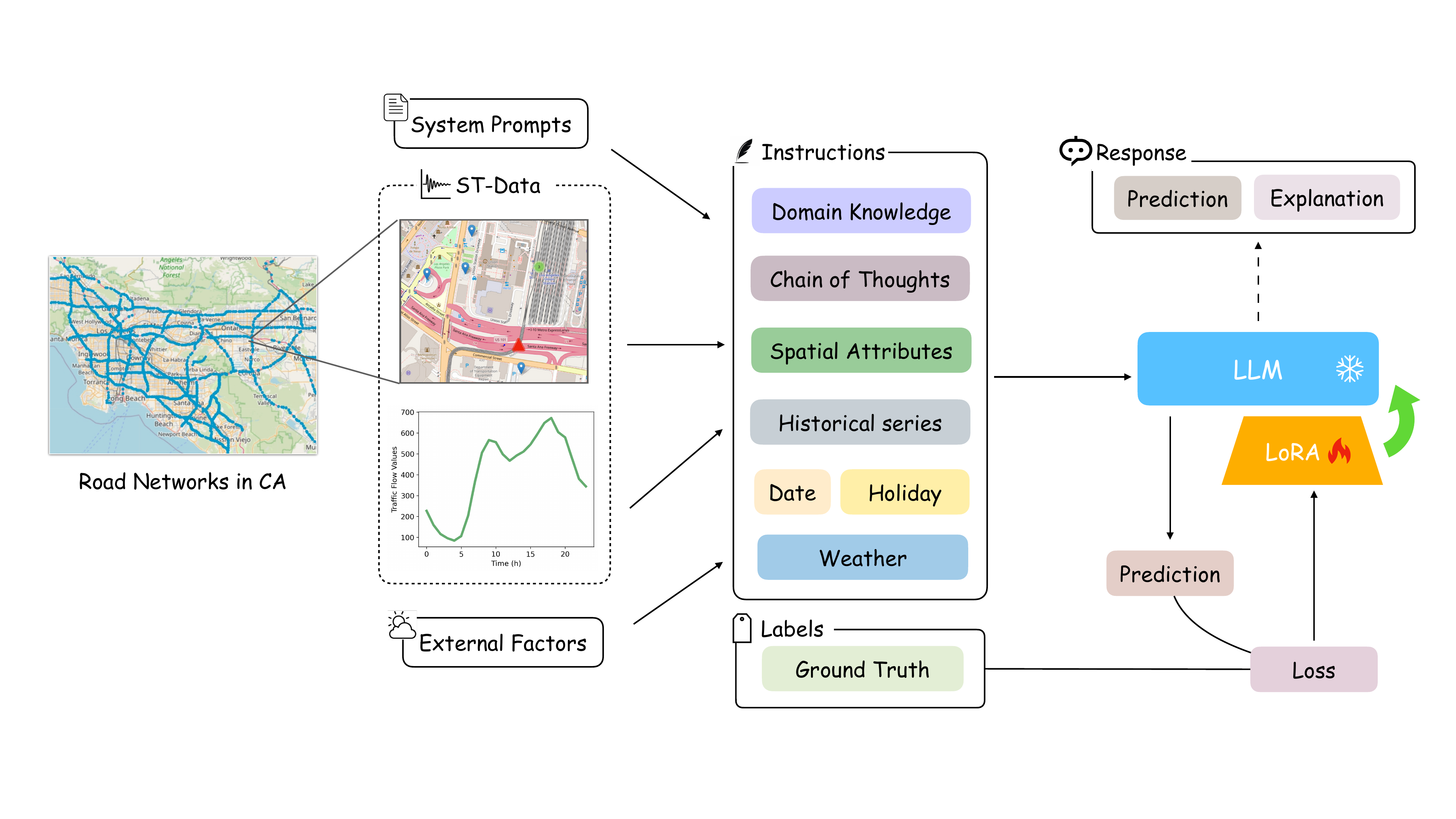}
        \caption{Framework of xTP-LLM: Multi-modal traffic flow data is converted into text-based prompts, leveraging task settings and domain knowledge to enable a Language Model (LLM) to discern latent relationships across diverse inputs and traffic patterns. Through fine-tuning, the LLM gains the ability to predict future values and furnish pertinent explanations. This refined model adeptly captures regional, traffic patterns and dependencies on input factors, facilitating generalized predictions even in unseen datasets, thus enhancing its predictive capabilities.}
\label{fig:framework}
\end{figure}
\begin{figure*}
    \centering
    \begin{minipage}{0.87\linewidth}
         \includegraphics[trim={8cm 2cm 10cm 2cm}, clip, width=\linewidth]{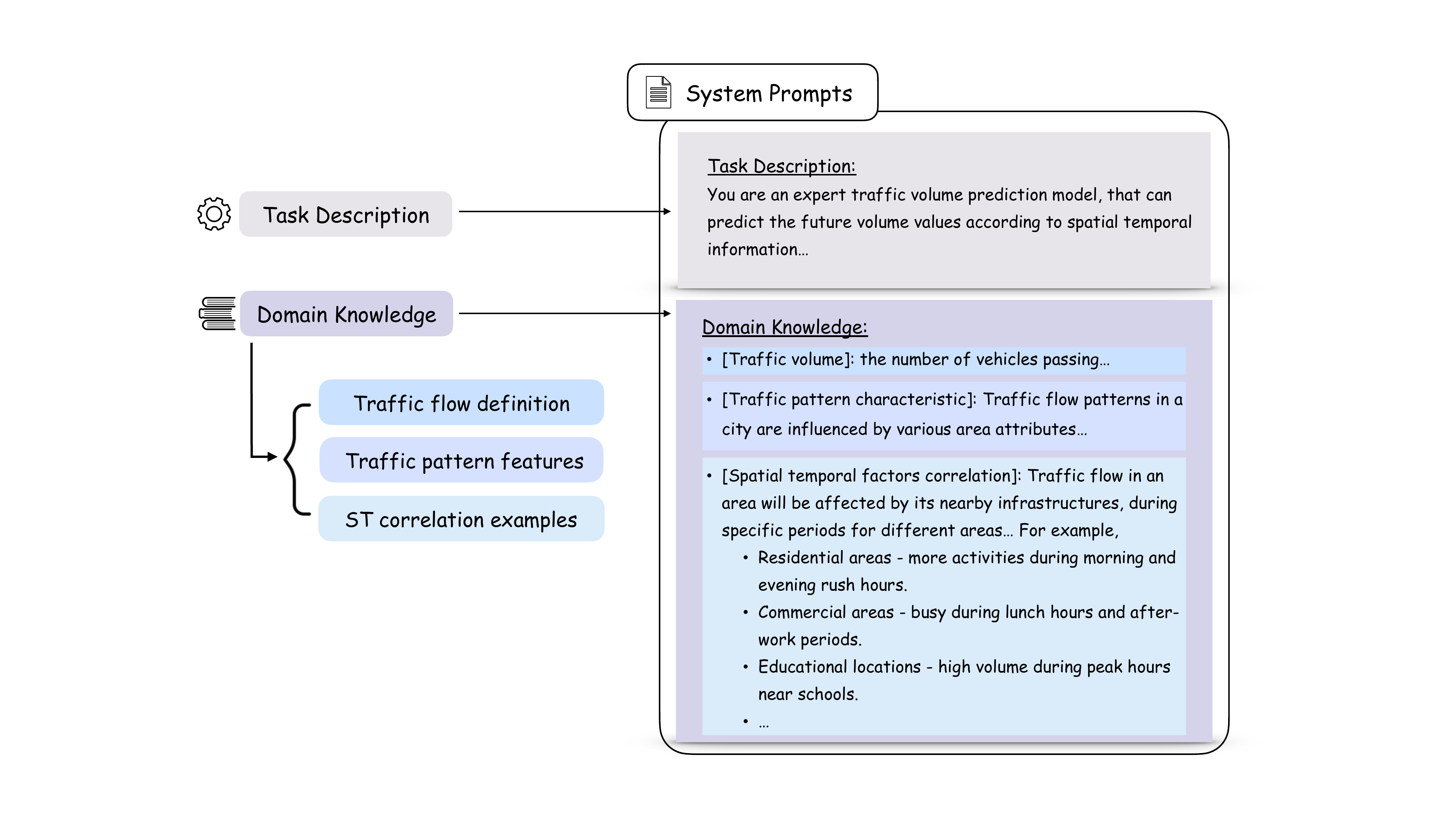}
        \subcaption{System prompts. They encompass task descriptions of traffic flow prediction and domain knowledge, including definitions, traffic pattern features, and examples of regional characteristics and traffic-pattern correlations. Equipped with these foundational settings and backgrounds, language models can tailor their knowledge and capabilities to the specific domain.}
        \label{fig:Ng1}
    \end{minipage}
    \begin{minipage}{0.87\linewidth}
         \includegraphics[trim={8cm 6cm 6cm 4cm}, clip, width=\linewidth]{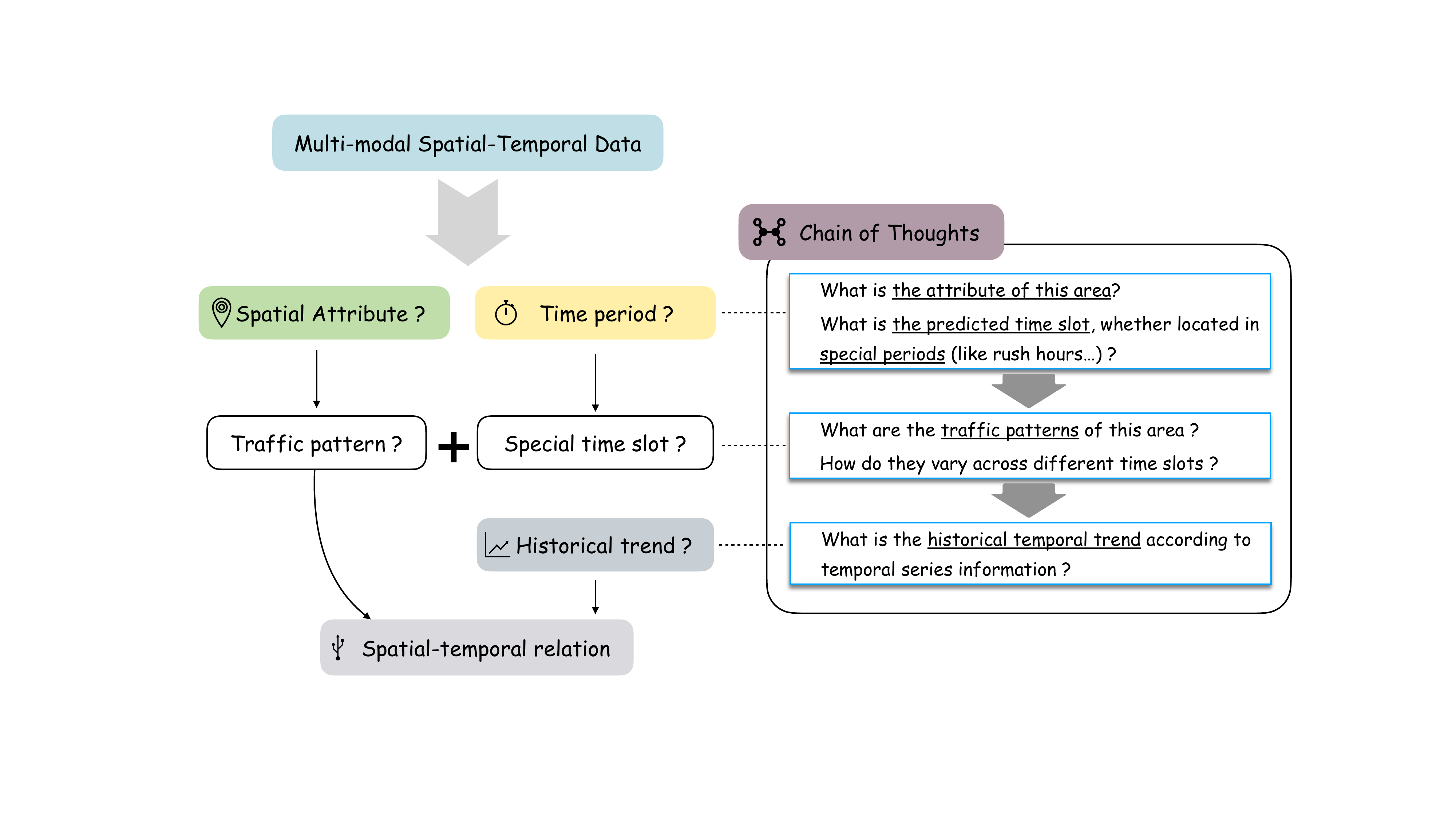}
        \subcaption{Chain of Thoughts. Chain of Thought (CoT) prompts encourage LLMs to extract relevant factors from provided information, prompting deeper consideration of their potential causal relationships with related knowledge. This process facilitates a comprehensive analysis and interpretation of the data.} 
        \label{fig:Ng2}
    \end{minipage}
  \caption{System Prompts and Chain of Thoughts construction.}
  \label{fig:sys_prompt}
\end{figure*}
\begin{figure*}
        \centering
        \includegraphics[trim={2cm 7cm 1cm 3cm}, clip,width=\linewidth]{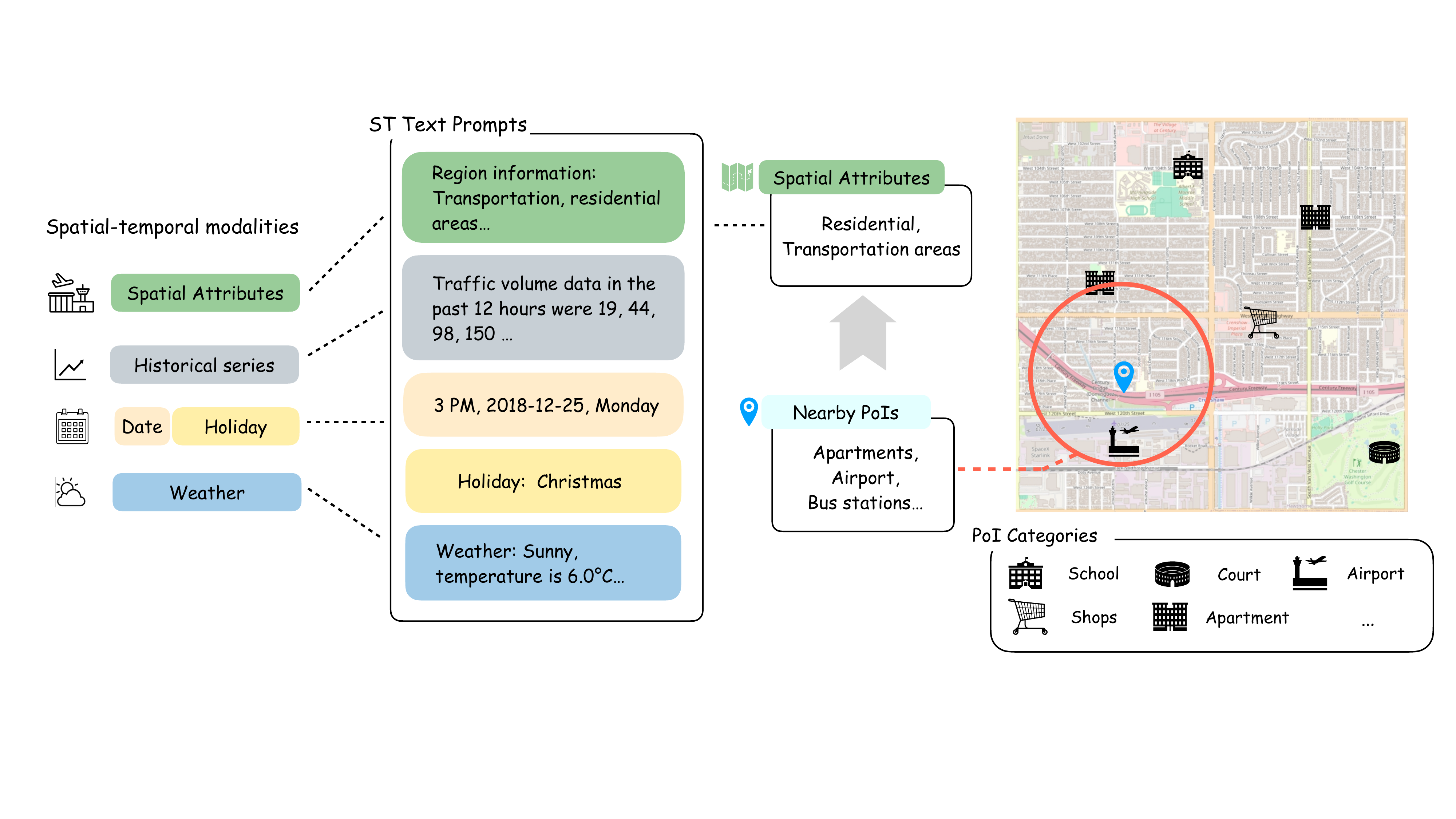}
        \caption{Multi-modal spatial-temporal(ST) text prompts. Spatial-temporal data covers spatial attributes and time series of historical traffic flow. To consider the spatial aspect of a region, nearby PoIs within a defined range are utilized to characterize its location features. For instance, consider a sensor positioned along the Century Freeway, surrounded by residential complexes and an airport, as illustrated in the above figure. This location can be classified as a blend of residential and transportation zones, reflecting the area's geographical attributes. To incorporate historical flow data and external factors, we represent these datasets textually as the rest components.}
\label{fig:st_prompt}
\end{figure*}

This section demonstrates how to textualize multi-modal traffic data and construct instructions to inspire the reasoning and predicting abilities of LLMs. The instruction part of Figure~\ref{fig:framework} showcases a meticulously crafted prompt template, designed to capture essential details consistently and comprehensively, including system prompts, spatial-temporal information, and external factors. Empirically, this structured format is tailored to convey diverse data modalities, boost the model's comprehension, and refine predictive accuracy.

\begin{itemize}
    \item \textbf{System Prompts}. To enhance the integration of LLMs into traffic prediction tasks, system prompts are structured with clear task descriptions and domain-specific knowledge, as shown in Figure~\ref{fig:Ng1}. Task settings explicitly outline the role of LLMs and leverage their pre-existing understanding of the traffic domain. Domain knowledge about traffic flow is embedded within the text description, incorporating few-shot examples to prompt LLMs to consider spatial factors, temporal fluctuation, and their interdependence. Spatial-specific traffic pattern cases are presented to assist models in establishing connections between geographical characteristics and traffic flow variations. For instance, residential areas may experience heightened traffic volume and increased travel activities during morning and evening rush hours.
    
    \item \textbf{Chain of Thoughts}. 
    Chain of Thought (CoT) \cite{wei2022chain} prompts have been proven to improve LLMs' reasoning capabilities in intricate problems, which promote models to think from shallow to deep. Inspired by zero-shot CoT \cite{kojima2022large}, spatial-temporal CoT prompts in Figure~\ref{fig:Ng2} are strategically crafted to enhance the LLMs' inference abilities. Initially, prompt questions guide LLMs to contemplate the spatial characteristics and potential traffic dynamics within a given area. Subsequently, LLMs assess whether the predicted time slot corresponds to special periods such as rush hours, weekdays, weekends, or holidays. After capturing spatial-temporal information, LLMs are directed to explore the profound connection between spatial data and the fluctuation of historical traffic flow, facilitating more precise predictions.
    
    \item \textbf{Spatial Attributes}. Multi-modal information prompts serve as the cornerstone of our approach. In this context, spatial attributes are derived from nearby Points of Interest (PoIs). We preprocess PoI category data within different proximity ranges (1km, 3km, 5km), aligning with the locations of traffic volume sensors. To reduce redundant input and avoid the LLM's over-reliance on PoI information, we summarized the PoI distribution into descriptions of regional attributes, such as transportation hubs, commercial zones, and residential areas, effectively representing the key characteristics of each geographical area. This approach facilitates the effective representation of spatial factors, enabling the model to grasp the intricate interplay between various factors impacting traffic flow dynamics. Also, regional data comprises details such as the city, and road location. Our xTP-LLM framework leverages this multifaceted information to recognize and integrate spatial and temporal patterns across diverse regions and periods. 
    
    \item \textbf{Historical time series}. Historical series are transcribed into textual descriptions with direct numerical representation. Data from the past 12 hours are illustrated for each hour's time slot, enabling LLMs to accurately perceive the temporal change of traffic flow, compared with encoding temporal features. This approach empowers LLMs to generate insights, understanding, and explanations for emerging trends and patterns.
    
    \item \textbf{External Factors}. External factors influencing traffic flow, such as dates, holidays, weather conditions, temperature, and vehicle visibility, are systematically considered. These diverse data points are uniformly transformed into textual information for comprehensive analysis.
\end{itemize}

Overall, this integration ensures a holistic understanding of the surrounding environment, historical trends, and external influences, enriching the model's contextual reasoning and enhancing its predictive capabilities.

\subsection{Supervised Fine-tuning}

Supervised fine-tuning is a key technique for adapting LLMs to specific tasks \cite{ding2023parameter, ziegler2019fine}. Initially, LLMs are trained on extensive, diverse datasets to develop broad linguistic knowledge. However, to excel in particular applications or domains, these models need additional training on task-specific data. In supervised fine-tuning, the pre-trained model is further trained on a dataset with labeled examples that are directly relevant to the target task. For instance, if the task is traffic prediction, the fine-tuning dataset would include historical traffic data paired with accurate forecasts. This process involves adjusting the model’s parameters to improve its performance by optimizing a loss function that measures the discrepancy between the model's predictions and the true outputs. Fine-tuning enables the model to leverage its general knowledge while becoming more proficient in the specific domain, enhancing both accuracy and relevance.

During the fine-tuning phase, the traffic forecasting task is framed as a next-token generation task.  The model predictions future traffic volumes autoregressively by optimizing the following objective function for the entire annotated dataset $\mathbb{D}$: 
\begin{align}
\quad \mathbb{L}^{FT}(\mathbb{D}) = -\sum_{j=1}^n \log P(Y_j|Y_{1:j-1}; X_{1:n}).
\end{align}
Here $Y_j$ is the ground truth of the next token, $X_{1:n}$ is the sequence of input historical tokens, and $Y_{1:j-1}$ is the previously generated tokens. Through this process, the model learns to make more accurate predictions tailored to the specific task or domain it is being fine-tuned for. This approach maximizes the learning potential of the LLM, enabling it to develop specialized proficiency in the target domain. 

Our xTP-LLM model is built upon the renowned open-source large language model, Llama2-7B-chat \cite{touvron2023llama}, and is fine-tuned with LoRA technique \cite{hu2022lora}. LoRA is a technique designed to efficiently adapt large pre-trained models to new tasks by introducing low-rank updates to the model's parameters. In LoRA, the adapted model's weight matrix $W$ is expressed as $W = W_0 + \alpha \Delta W$. where $W_0$ represents the original pre-trained weights, $\alpha$ is the scaling factor that determines the magnitude of the low-rank update, and $\Delta W$ is the update applied during fine-tuning. Instead of updating the entire matrix $\Delta W$ directly, LoRA approximates it using the product of two lower-dimensional matrices: $\Delta W = B \times A$. Here, $B \in \mathbb{R}^{d \times r}$ and $A \in \mathbb{R}^{r \times k}$, with $r$ being the rank of the approximation, chosen to be much smaller than the dimensions of $W_0$ (i.e. $r \ll \min(d, k)$). This low-rank factorization significantly reduces the number of parameters involved in the update from $d \times k$ to $ r \times (d + k)$, leading to substantial computational and memory savings.

\subsection{Explanations Generation}
The chat models in Llama2 are tailored to excel in understanding and generating text in conversational contexts. Therefore, by incorporating explanation requirements into the prompts, our model can not only generate prediction results but also provide explanations simultaneously. Initially, we add instructions for generating explanatory text directly within the input prompt. Although xTP-LLM was able to produce such explanations, they often lacked coherence with the prediction results. We attribute this issue to the model's insufficient alignment between explanations and predictions, as only the predictions are used in loss calculations during the fine-tuning phase. We addressed this misalignment by employing few-shot learning \cite{radford2019language, dai2020nonlinear}. By including a few carefully selected examples in the input prompt—examples that illustrate how the explanation should align with the traffic flow prediction—the LLM is able to learn this alignment dynamically during inference. To conduct few-shot learning in our context, we follow the steps below:
\begin{itemize}
    \item \textbf{Create Few-Shot Examples:} We selected 2-3 examples where the traffic flow prediction and the corresponding explanations were correctly aligned. These examples are generated through ChatGPT with ground truth predictions in input. 
    \item \textbf{Incorporate into the Input Prompt:} These few-shot examples were then included directly in the input before the instruction fine-tuning prompt.
    \item \textbf{Model Inference:} The LLM uses these examples as a guide during inference, enabling it to generate explanations that align well with the predicted sequences.
\end{itemize}
By structuring the few-shot examples in this way, we effectively leveraged the LLM's in-context learning capability to resolve the misalignment issue without requiring extensive new data or additional fine-tuning. This approach not only improved the coherence of the output explanations but also demonstrated the adaptability of LLMs in handling complex tasks with minimal examples.

\section{Experimental Settings and Results}
In this section, we will start by introducing our experimental setups, which include the dataset we used, the evaluation methods we employed, the baseline models we compared against, and the parameter settings we used in the fine-tuning process. Subsequently, we will present the main results of our experiments, which involve comparing xTP-LLM with the baseline models, discussing how the performance of our model varied in the spatio-temporal domain, conducting ablation studies, and exploring the generalization capabilities of our model. Finally, we will discuss the explanations generated by xTP-LLM in corresponding to the traffic flow predictions. This discussion will elucidate how the robust reasoning capability inherent in large language models enhances traffic forecasting tasks.

\subsection{Dataset Description}
Our experiments were conducted based on our proposed multi-modal traffic prediction dataset, named CATraffic. This dataset consists of traffic volume data from various regions in California, as well as meteorological information, nearby point of interests (PoIs) data, and holiday information. The traffic volume data is sourced from the LargeST dataset \cite{liu2023largest}, which comprises five years (2017-2021) of traffic flow data in California, encompassing 8600 traffic sensors sampled at a 15-minute interval. We constructed CATraffic by selecting a subset of the LargeST dataset, focusing on 1000 sensors from the Greater Los Angeles (GLA) and Greater Bay Area (GBA). Our dataset spans two years from January 1, 2018, to December 30, 2019, with data sampled hourly. The choice of a 1-hour granularity represents a strategic compromise that balances the need for temporal resolution with the computational constraints as well as the input context length constraints of large language models. Our PoIs data is obtained through \textit{OpenStreetMap}\footnote{OpenStreetMap: https://openmaptiles.org} with the help of the \textit{Overpass Turbo API}\footnote{Overpass Turbo: https://overpass-turbo.eu}. For the meteorological data, we collected information from the National Oceanic and Atmospheric Administration (\textit{NOAA}\footnote{NOAA: https://www.ncei.noaa.gov/}). These include factors such as reported weather events, temperature, and visibility, which are considered as they have direct impacts on traffic patterns.

During the sensor selection process, to ensure a diverse representation of traffic patterns, we clustered all 8600 sensors into 1000 categories based on their nearby PoIs features. These PoIs features are formalized as $\mathbb{F}_{poi} = [(f^e_1, f^e_2, ..., f^e_n), (f^w_1, f^w_2, ..., f^w_n), (f^n_1, f^n_2, ..., f^n_n), (f^s_1, f^s_2, ..., f^s_n)]$, where $f^e$, $f^w$, $f^n$, and $f^s$ represent the PoIs features in the four directions (East, West, North, and South). We selected the top-$n$ PoI categories, normalized their counts in a range of 5 km, and expressed them as $f_i$ for $i$-th category. In our experiments, we set n to 20 for convenience. These PoIs feature vectors were then used as input for clustering via the K-means algorithm, which grouped the sensors into 1000 clusters. For each cluster, we selected one representative sensor, resulting in a final set of 1000 sensors. This approach effectively reduces the dataset to a manageable size while preserving a diverse representation of the various spatial and feature characteristics present in the original dataset.

We split the collected data into two parts: data from 2018 was used as the training set, while data from 2019 was reserved for model validation. We believe an entire year of data for testing provides a more comprehensive assessment of the model’s performance across a full spectrum of temporal variations, including different weather conditions and holiday periods. All experiments were configured to predict traffic flows for the next 12 hours based on historical 12-hour traffic flow data. During the data preprocessing stage, we filtered out samples with zero values for 24 consecutive hours, which likely resulted from malfunctioning sensors. Additionally, to assess the model's generalization capability, we created a zero-shot dataset derived from LargeST \cite{liu2023largest}. This dataset comprises data from 100 sensors (also selected through clustering) in San Diego (SD), covering the period from November 1, 2019, to December 31, 2019. These data were not utilized in the model fine-tuning process and served to the evaluation of the model's generalization performance. 

\subsection{Evaluation Metrics}
In time series forecasting tasks, researchers commonly employ Root Mean Square Error (RMSE), Mean Absolute Error (MAE), and Mean Absolute Percentage Error (MAPE), to evaluate the accuracy of forecasting results. These metrics are defined as follows:
\begin{align}
RMSE =& \sqrt{\frac{1}{n}\sum_{i=1}^{n}(y_i - \hat{y}_i)^2} \\
MAE =& \frac{1}{n}\sum_{i=1}^{n}|y_i - \hat{y}_i| \\
MAPE =& \frac{1}{n}\sum_{i=1}^{n}\left|\frac{y_i - \hat{y}_i}{y_i}\right| \times 100\%
\end{align}
Here, $y_i$ represents the ground truth value of the $i$-th data point, $\hat{y}_i$ denotes the corresponding prediction value, and $n$ stands for the total number of samples. RMSE is advantageous for emphasizing larger errors due to its square term, while MAE provides a straightforward interpretation by averaging absolute errors, treating all errors equally. On the other hand, MAPE measures the average percentage difference between predictions and ground truths, offering interpretability in terms of relative accuracy but being sensitive to zero values in the denominator. Researchers typically employ a combination of these metrics to comprehensively assess model performance.

\subsection{Baseline Models}
We extensively compared our proposed xTP-LLM with 9 advanced baseline models. Among these, LSTM \cite{graves2012long} stands as a temporal-only deep model based on Recurrent Neural Networks (RNNs), disregarding spatial correlations. Additionally, we select DCRNN \cite{li2017diffusion} and AGCRN \cite{bai2020adaptive} as the representation of RNNs-based methods. We also choose TCN-based methods such as STGCN \cite{yu2017spatio} and GWNET \cite{wu2019graph}, along with attention-based methods ASTGCN \cite{guo2019attention} and STTN \cite{xu2020spatial}. These models were proposed between 2018 and 2020, reflecting the prevalent research direction in time series forecasting during those years. Furthermore, we integrated three representative methods from recent years, including STGODE \cite{fang2021spatial} and DSTAGNN \cite{lan2022dstagnn}. STGODE adeptly utilizes neural ordinary differential equations to capture the continuous changes of traffic signals, while DSTAGNN is specifically designed to capture the dynamic correlations among traffic sensors.

\subsection{Experiment Settings}
Our model is fine-tuned based on the well-known open-source large language model, Llama2 \cite{touvron2023llama}, specifically utilizing the chat version with a size of 7B. We load the base model in 8 bits for fine-tuning, with training parameters including a batch size of 8, a learning rate of 5e-4, a warm-up step of 400, gradient accumulation steps of 8, and a training epoch for 2. The LoRA parameters are configured with a rank of 16 and an alpha value of 32. During the inference phase, a temperature of 0.95 was applied. 

\subsection{Overall Performance}

\begin{table}[!h]
\scriptsize
\begin{tabularx}{\textwidth}{@{} c | l | *{10}{c} @{}} 
\toprule 
\centering Steps & Metrics & LSTM & DCRNN & STGCN & ASTGCN & GWNET & AGCRN & STTN & STGODE & DSTAGNN & xTP-LLM \\
\midrule
\multirow{3}{*}{3} & RMSE & 62.65 & 61.12 & 43.43 & 66.60 & \underline{42.59} & 43.92 & 43.41 & 52.30 & 51.79 & \textbf{40.34} \\
 & MAE & 40.00 & 37.32 & 25.83 & 44.47 & \underline{25.56} & 25.67 & 26.00 & 29.53 & 34.06 & \textbf{20.20} \\
 & MAPE(\%) & 24.75 & 24.52 & 16.09 & 28.36 & 15.96 & \underline{15.60} & 16.22 & 19.69 & 23.90 & \textbf{9.94} \\
\midrule
\multirow{3}{*}{6} & RMSE & 74.69 & 74.21 & 46.64 & 75.07 & \underline{45.91} & 48.92 & 51.63 & 55.79 & 53.44 & \textbf{44.53} \\
 & MAE & 48.10 & 45.83 & \underline{27.94} & 49.47 & 28.27 & 28.94 & 30.79 & 32.97 & 33.72 & \textbf{22.70} \\
 & MAPE(\%)  & 33.98 & 31.58 & \underline{15.75} & 31.10 & 18.00 & 17.31 & 18.99 & 21.62 & 21.27 & \textbf{11.39} \\
\midrule
\multirow{3}{*}{9} & RMSE & 77.61 & 78.22 & 50.04 & 81.76 & \underline{46.61} & 52.06 & 56.19 & 59.97 & 56.21 & \textbf{45.89} \\
 & MAE & 49.60 & 47.97 & 30.16 & 53.43 & \underline{28.70} & 30.74 & 33.11 & 35.23 & 36.45 & \textbf{23.89} \\
 & MAPE(\%)  & 39.26 & 32.22 & \underline{17.52} & 35.35 & 19.88 & 18.38 & 20.69 & 22.44 & 24.00 & \textbf{12.09} \\
\midrule
\multirow{3}{*}{12} & RMSE & 69.86 & 71.79 & 54.91 & 69.75 & \underline{49.79} & 56.13 & 61.38 & 60.47 & 60.33 & \textbf{47.82} \\
 & MAE & 42.05 & 43.08 & 33.20 & 43.40 & \underline{30.47} & 32.70 & 38.27 & 36.22 & 37.91 & \textbf{24.99} \\
 & MAPE(\%)  & 25.99 & 25.05 & \underline{19.27} & 28.03 & 20.46 & 19.50 & 26.09 & 23.16 & 24.23 & \textbf{12.30} \\
\midrule
\multirow{3}{*}{Avg.} & RMSE & 68.14 & 67.89 & 46.69 & 70.48 & \underline{43.97} & 47.97 & 50.03 & 54.41 & 52.82 & \textbf{42.81} \\
 & MAE & 43.23 & 41.45 & 28.02 & 46.17 & \underline{26.84} & 28.22 & 30.04 & 31.95 & 34.02 & \textbf{21.91} \\
 & MAPE(\%)  & 30.21 & 27.14 & \underline{16.68} & 29.99 & 17.44 & 16.94 & 19.04 & 20.72 & 22.45 & \textbf{11.01} \\
\bottomrule
\end{tabularx}
\caption{Performance comparison between our proposed xTP-LLM and baseline models on our CATraffic dataset. The horizon for traffic flow prediction is 12, and we show the results in Steps 3, 6, 9, and 12 in the table. The last three rows are the average performance for all 12 steps. We highlight the best results in bold and the second best in underline.} 
\label{tab:result_comp}
\end{table}

\begin{figure}[!h]
    \centering
    \includegraphics[width=\linewidth]{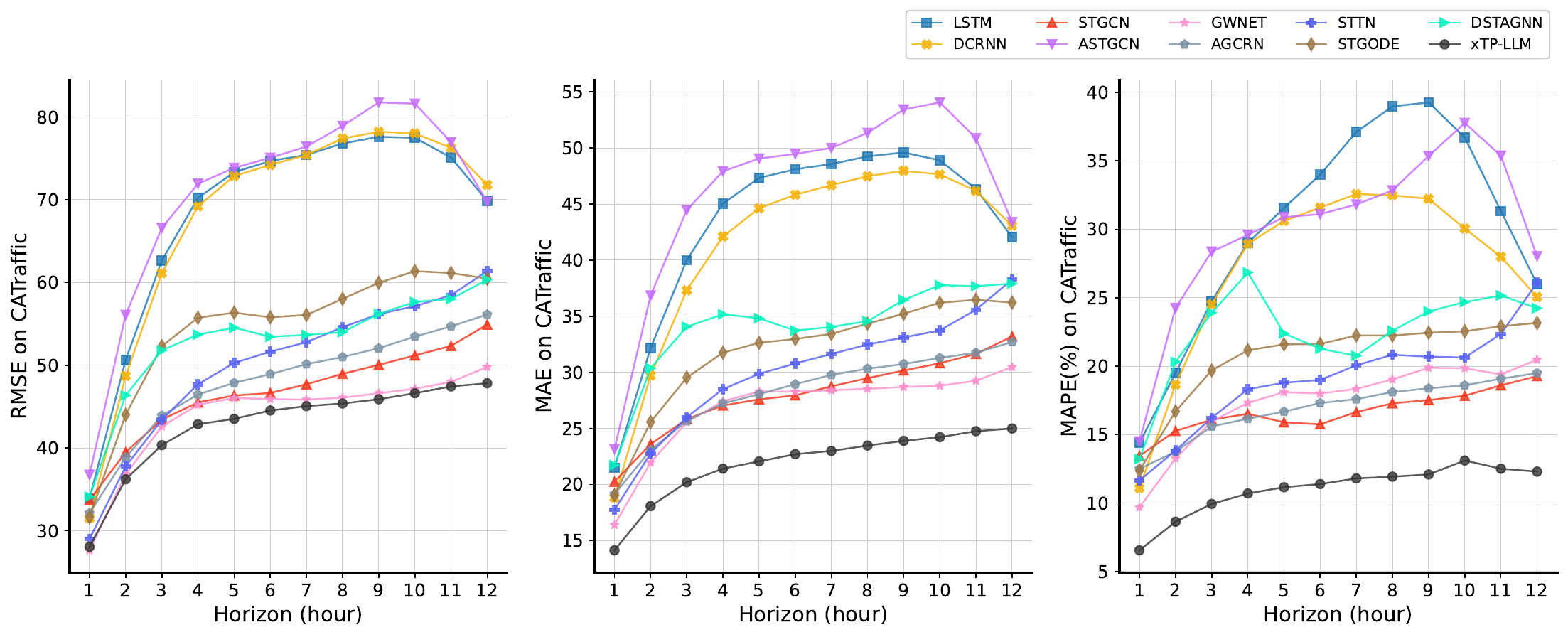}
    \caption{Time step-based prediction comparison with baseline models and our xTP-LLM. All experiments are conducted based on our CATraffic dataset. Results show that the performance of all models decreases with increasing prediction time, whereas our model significantly outperforms the others in all prediction steps.}
\label{fig:baseline_comp}
\end{figure}

We report the performance comparison results between xTP-LLM and the baseline models in Table~\ref{tab:result_comp}. All models are trained and evaluated based on our CATraffic dataset, with the same dataset settings. The task is to utilize traffic flow data from the historical 12 hours to forecast future traffic flows in the next 12 hours. We represent the results of horizons 3, 6, 9, and 12, as well as the average performance over all 12 steps in the table. The results demonstrate that the overall performance of our proposed xTP-LLM exceeds that of the baseline models by a large margin, especially in the MAE and MAPE. For example, In terms of average performance of all 12 horizons, our model outperforms the best two baseline models, GWNET \cite{wu2019graph} by 18.37\% in MAE, and STGCN \cite{yu2017spatio} by 34.00\% in MAPE, which shows the impressive capability in traffic flow forecasting of our model.

We further demonstrate the results with different prediction horizons of the baseline models and xTP-LLM, depicted in Figure~\ref{fig:baseline_comp}. From left to right, RMSE, MAE, and MAPE of compared models at different prediction steps are shown. The results yielded the following observations:  
\begin{itemize}
    \item As the prediction horizon increases, performance generally declines across all models, as longer-term forecasts inherently entail greater uncertainty and complexity. However, several models exhibit improved performance in longer-term forecasting, such as LSTM, ASTGCN, and DCRNN. This phenomenon may be attributed to these models' ability to capture and leverage the periodicity within the data, allowing them to make more accurate predictions over extended time horizons.
    \item Our proposed model consistently outperforms the comparison methods at each time step, showing significant advantages in both short-term and long-term traffic flow forecasting. This indicates the robustness of our model in various prediction horizons.
    \item Our proposed model shows a more significant advantage in MAE and MAPE than RMSE, which may be attributed to the amplification effect of RMSE on outliers, leading to inaccurate assessments. Compared to RMSE, MAE and MAPE are less sensitive to extreme errors because they measure the mean absolute error and the percentage error, respectively. The excellent performance of our model on MAE and MAPE suggests that it is effective in mitigating the effects of outliers and provides more accurate and stable predictions, especially in cases where extreme values may occasionally occur.
\end{itemize}

These findings underscore the effectiveness of our method in capturing complex temporal patterns in traffic flow data, leading to more accurate and reliable predictions.

\subsection{Spatial and Temporal Homogeneity}

Evaluating spatial and temporal homogeneity in traffic prediction is vital for evaluating model performance, generalizability, and robustness. It ensures that traffic flow prediction models can effectively adapt to diverse real-world conditions. Therefore, in this section, we will thoroughly analyze the performance of our proposed model in terms of spatio-temporal consistency.

\begin{figure}[!h]
    \centering
    \begin{minipage}{0.58\linewidth}
        \includegraphics[width=\linewidth]{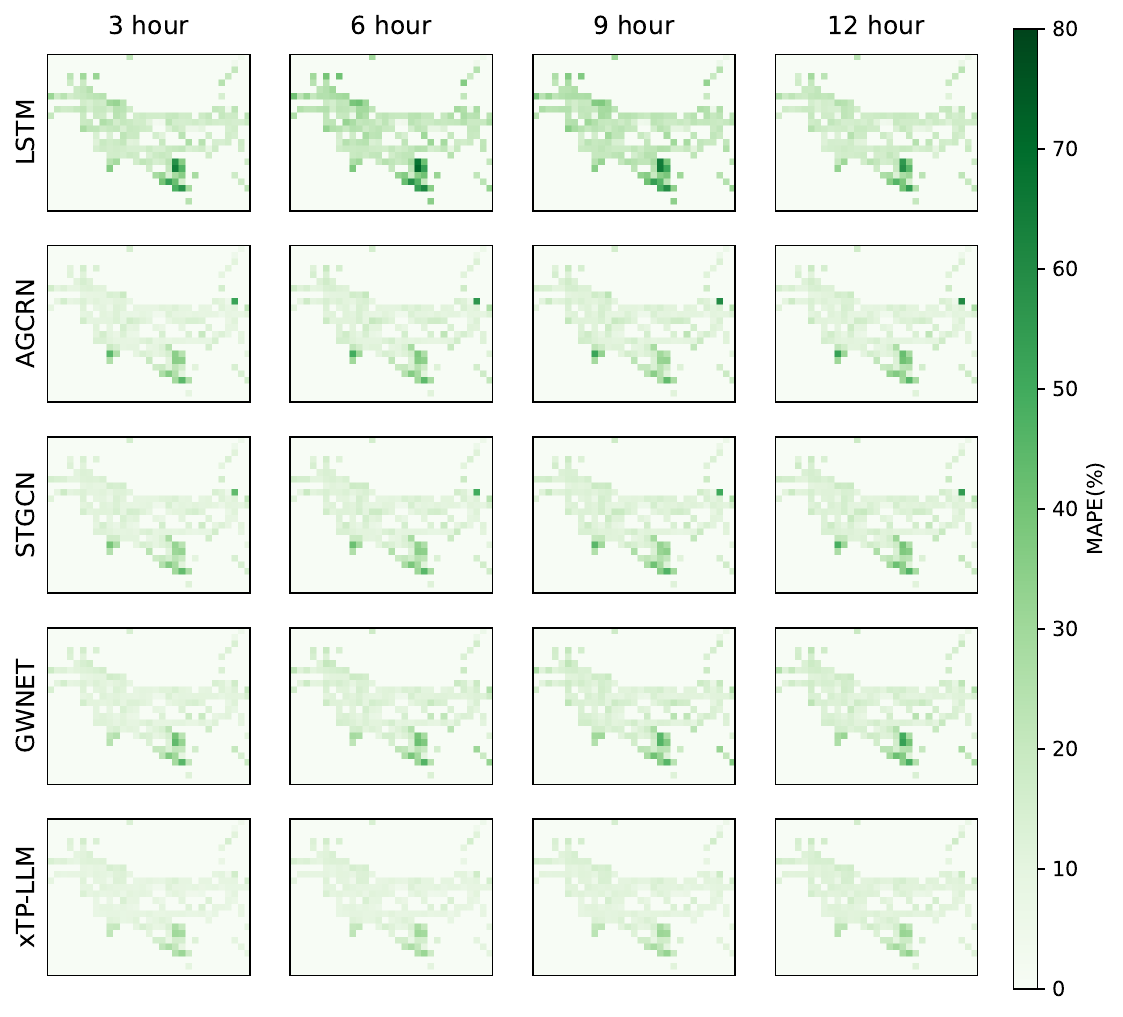}
        \caption*{(a) Spatial MAPE map}
        \label{fig:la_error_map}
    \end{minipage}
    \begin{minipage}{0.58\linewidth}
        \includegraphics[width=\linewidth]{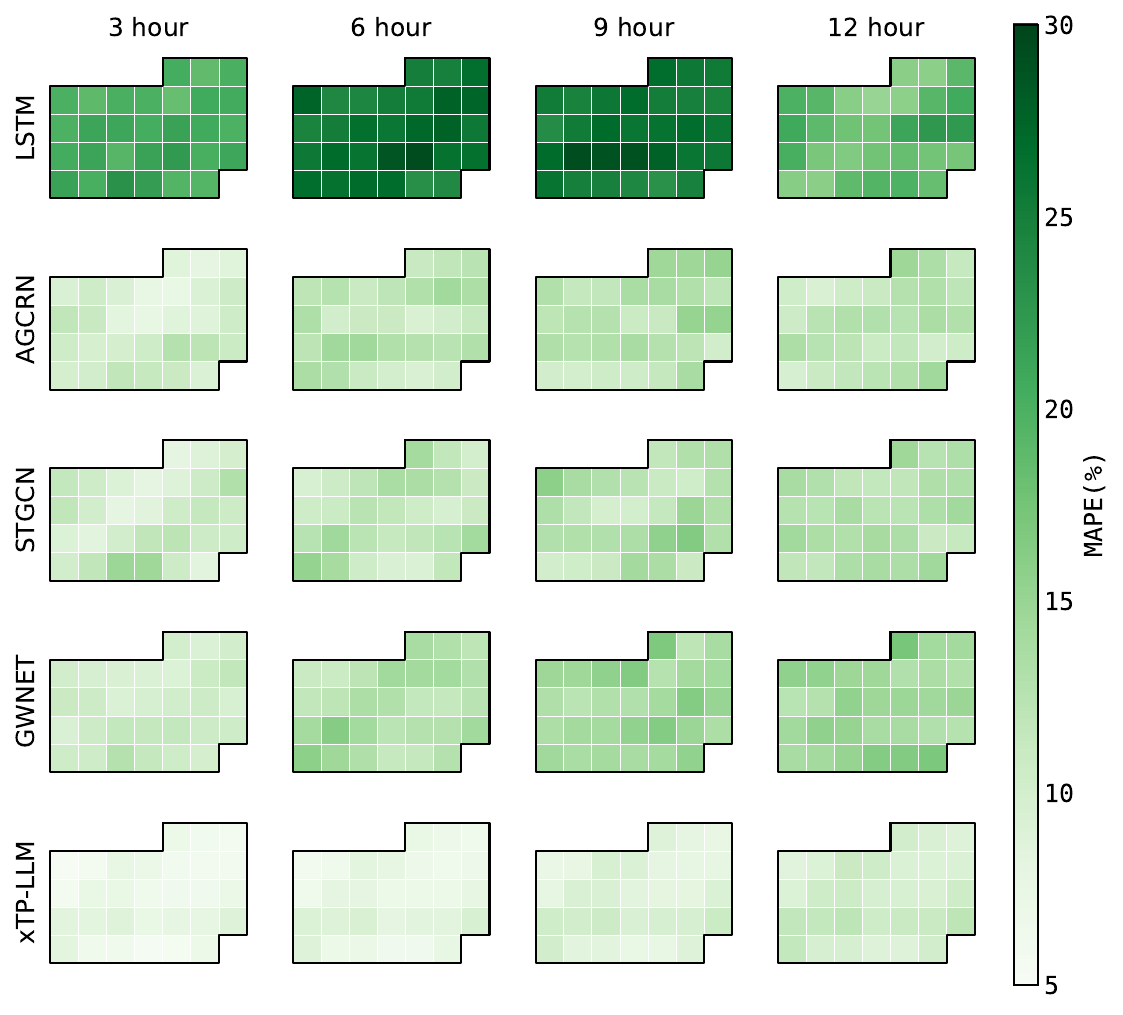}
        \caption*{(b) Temporal MAPE map}
        \label{fig:month_error_map}
    \end{minipage}
  \caption{Comparison of the spatial and temporal prediction error distributions among different models. Figure (a) presents the distribution of MAPE for our proposed xTP-LLM and four baseline models in various locations across the Greater Los Angeles (GLA), considering four different prediction horizons. Figure (b) illustrates the distribution of daily average MAPE over the whole November in 2019, with four different prediction horizons.}
  \label{fig:st_homo}
\end{figure}

Spatial homogeneity helps to assess the ability of different models to learn traffic patterns at different spatial locations. To evaluate this kind of capability, we tested our xTP-LLM as well as four baseline models (LSTM, AGCRN, STGCN and GWNET) at different locations with varying urban characteristics in the Greater Los Angeles Area (GLA). The result, depicted in Figure~\ref{fig:st_homo} (a), illustrates MAPE values for four prediction horizons of 3h, 6h, 9h, and 12h, with darker colors indicating poorer performance. Overall, our xTP-LLM model exhibits relatively consistent prediction performance across different locations, effectively capturing traffic flow trends irrespective of varied spatial characteristics. In contrast, other models demonstrate inferior homogeneity, particularly in areas featuring complex road network intersections and intricate facility distributions. These findings underscore the robustness and adaptability of our proposed xTP-LLM model in effectively learning and predicting traffic patterns across diverse spatial contexts, thus highlighting its potential for real-world application in urban traffic management and planning scenarios.

In the analysis of temporal homogeneity, we evaluated models on a subset of the test dataset (traffic data in November 2019) and reported the daily averaged MAPE values. Illustrated in Figure~\ref{fig:st_homo} (b) as calendar heat maps, the results showcase the comparative performance across the month. Our model consistently exhibits lower daily average MAPE, demonstrating robustness in capturing temporal nuances of traffic flow patterns. Notably, it consistently outperforms others throughout the whole month, emphasizing its superior capability in handling diverse temporal dynamics inherent in real-world traffic scenarios.

\subsection{Ablation Studies}
\subsubsection{The impact of input components on model performance}

\begin{table}[!h]
\footnotesize
\centering
  \begin{tabular}{c|ccc|cc|ccc}
\toprule 
\centering Input Settings & Date & Weather & PoIs & Domain Knowledge & CoT Prompting & RMSE & MAE & MAPE(\%) \\
\midrule
A & \XSolidBrush & \XSolidBrush & \XSolidBrush & \XSolidBrush & \XSolidBrush & 89.17 & 49.49 & 30.08\\
B & \textcolor{red}{\Checkmark} & \XSolidBrush & \XSolidBrush & \XSolidBrush & \XSolidBrush & 56.43 & 30.87 & 17.05\\
C & \XSolidBrush & \textcolor{red}{\Checkmark} & \XSolidBrush & \XSolidBrush & \XSolidBrush & 82.43 & 47.25 & 28.52\\
D & \XSolidBrush & \XSolidBrush & \textcolor{red}{\Checkmark} & \XSolidBrush & \XSolidBrush & 75.14 & 42.28 & 26.39\\
E & \textcolor{red}{\Checkmark} & \textcolor{red}{\Checkmark} & \XSolidBrush & \XSolidBrush & \XSolidBrush & 51.51 & 28.35 & 14.98\\
F & \textcolor{red}{\Checkmark} & \XSolidBrush & \textcolor{red}{\Checkmark} & \XSolidBrush & \XSolidBrush & 48.50 & 24.31 & 12.88 \\
G & \XSolidBrush & \textcolor{red}{\Checkmark} & \textcolor{red}{\Checkmark} & \XSolidBrush & \XSolidBrush & 69.97 & 40.44 & 24.90\\
H & \textcolor{red}{\Checkmark} & \textcolor{red}{\Checkmark} & \textcolor{red}{\Checkmark} & \XSolidBrush & \XSolidBrush & 46.20 & 22.36 & 12.05 \\
I & \textcolor{red}{\Checkmark} & \textcolor{red}{\Checkmark} & \textcolor{red}{\Checkmark} & \textcolor{red}{\Checkmark} & \XSolidBrush & 44.22 & 22.64 & 11.23 \\
J & \textcolor{red}{\Checkmark} & \textcolor{red}{\Checkmark} & \textcolor{red}{\Checkmark} & \XSolidBrush & \textcolor{red}{\Checkmark} & 45.49 & 23.55 & 12.38 \\
K & \textcolor{red}{\Checkmark} & \textcolor{red}{\Checkmark} & \textcolor{red}{\Checkmark} & \textcolor{red}{\Checkmark} & \textcolor{red}{\Checkmark} & \textbf{42.81} & \textbf{21.91} & \textbf{11.01} \\
\bottomrule
  \end{tabular}
  \caption{Ablation study results showing the impact of different input settings (Date, Weather, PoIs, Domain Knowledge, and CoT Prompting) on model performance.}
  \label{tab:ablation_study}
\end{table}

To assess the effect of varying prompt input on model performance, we conducted a thorough ablation analysis. This approach enabled us to systematically evaluate the contribution of each component to the overall performance of the model. Specifically, we investigated the effects of including date, weather, PoIs information, domain knowledge, and CoT prompting on model performance, and the results presented in Tab.~\ref{tab:ablation_study}. We observed that date, weather, and points of interest (PoIs) each independently enhance model performance, as evidenced by comparisons between input settings B, C, D, and A. Notably, date information yields the most significant improvement, with a 37.62\% reduction in MAE, underscoring its critical role in enhancing predictive accuracy. Weather and PoIs also contribute positively, with MAE reductions of 4.53\% and 14.58\%, respectively, though their impact is less pronounced compared to date information. Moreover, The interaction between date, weather, and points of interest (PoIs) significantly enhances model performance, as demonstrated by input settings E, F, and G. Any combination of these three components can further improve the model's predictive performance, illustrating their synergistic effect. For example, setting F results in an MAE of 24.31, shows a 21.25\% improvement compared to setting B, which shows the synergistic effect of date and PoIs on model performance. Similarly, the including weather and PoIs (setting G) shows a 14.41\% improvement in MAE compared to setting C.  These results underscore how the integration of date, weather, and PoIs interactively boosts model performance, with each component amplifying the benefits provided by the others. 

Additionally, an interesting observation emerges when comparing input settings H, I, J, and K. We found that adding domain knowledge or CoT prompting alone did not significantly improve model performance and sometimes even degraded it (as seen in settings I and J compared to H). However, when both were combined (setting K), significant performance improvements were observed, with a 4.57\% reduction in MAE. This finding contrasts with our initial expectations and suggests that domain knowledge and CoT prompting work synergistically. We attribute this to CoT prompting effectively guiding the model in utilizing domain knowledge, extracting relevant information for the prediction task. Conversely, CoT prompting also benefits from the inclusion of domain knowledge, enabling it to achieve its full potential.

\subsubsection{The impact of different horizons on model performance}

\begin{figure}[t]
        \centering
        \includegraphics[width=0.8\linewidth]{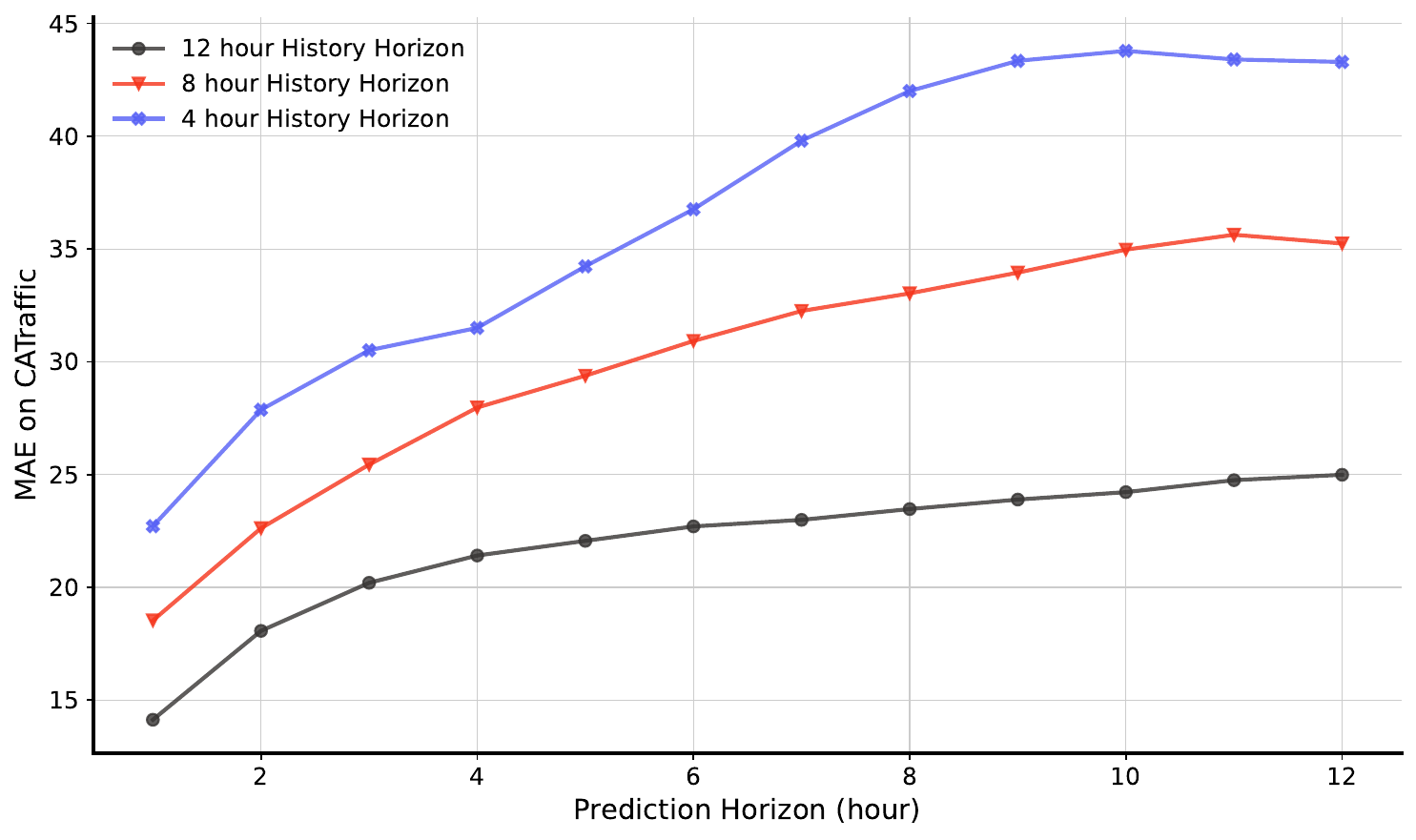}
        \caption{Model performances with different history horizons. We show that as the history horizon shortens, the MAE increases significantly, indicating that the model performs better with longer historical data.}
\label{fig:diff_his_horizon}
\end{figure}

To explore the impact of different horizons on model performance, we conducted additional experiments with historical horizons of 4 and 8 hours. The evaluation results in the test dataset are presented in the Fig.~\ref{fig:diff_his_horizon}. We observed that as the historical horizon decreases, the model's prediction error, measured by MAE, increases significantly. Specifically, the 12-hour horizon consistently yielded the lowest MAE values, indicating superior performance compared to shorter horizons. In contrast, the 4-hour horizon exhibited the highest MAE, suggesting that a shorter historical horizon substantially impairs the model's ability to accurately predict traffic flow. This is because a longer historical horizon provides more information, allowing the model to more accurately identify traffic flow patterns. Moreover, examining different prediction steps provides a direct view of how varying prediction horizons impact the model’s predictions. As expected, the performance deteriorates as the prediction horizon increases in all three historical horizons. Two primary factors contribute to this phenomenon. Firstly, the increasing temporal distance introduces greater uncertainty and variability into the forecasting process, thereby exacerbating the challenge of accurately predicting future events. Secondly, the iterative nature of large language model (function as a next-token predictor) means that any error in early prediction is propagated and magnified in subsequent steps, leading to a cumulative effect that significantly impairs the model's overall performance on the long range.

\subsection{Generalization Studies}

\subsubsection{Robustness analysis on the test dataset}

\begin{table}[!h]
\centering
\begin{tabular}{c|m{3.2cm}|m{3.2cm}|m{1.5cm}|m{1.5cm}}
\toprule
Different periods & Peak hours \newline (7 AM - 9 AM, \newline 4 PM - 6 PM) &  Off-peak hours \newline (10 AM - 3 PM, \newline 7 PM - 7 AM) & weekdays & weekends \\
\midrule
Avg. MAE & \centering 21.25 & \centering 22.57 & \centering 22.05 & \quad 21.56 \\
\bottomrule
\end{tabular}
\caption{Model's performance across different time periods, including peak and off-peak hours, weekdays and weekends.}
\label{tab:diff_time_eval}
\end{table}

\begin{table}[!h]
\centering
\begin{tabular}{m{2.8cm}|ccccccc}
\toprule
Weathers & Sunny & Rain & Foggy & Thunderstorm & Sleet & Storm  & Snow  \\
\midrule
Percentages on training dataset & 83.3\% & 11.6\% & 4.7\% & 0.43\% & 0.016\% & 0.0\% & 0.0\% \\

Percentages on test dataset & 76.4\% & 19.8\% & 3.0\% & 0.74\% & 0.0\% & 0.028\% & 0.011\% \\

Avg. MAE & 21.88 & 21.42 & 20.05 & 22.11 & - & 20.17 & 19.93 \\
\bottomrule
\end{tabular}
\caption{The table shows the distribution of various weather conditions in the training and test datasets, as well as the model's performance across these different weather conditions.}
\label{tab:diff_weather_eval}
\end{table}

To ensure the robustness of our xTP-LLM model, we conducted a comprehensive analysis of its performance on the test dataset, focusing on various time periods such as peak hours, off-peak hours, weekdays, weekends, and different weather scenarios. The results, detailed in Tab.~\ref{tab:diff_time_eval} and Tab.~\ref{tab:diff_weather_eval}, demonstrate the model's resilience and consistency across these diverse scenarios.

Tab.~\ref{tab:diff_time_eval} illustrates the model's performance across different time periods, including peak hours, off-peak hours, weekdays, and weekends. The average MAE values range from 21.25 to 22.57, indicating that the model maintains a stable level of accuracy regardless of temporal variations. This consistency underscores the robustness of the xTP-LLM model in handling different time-related scenarios. Tab.~\ref{tab:diff_weather_eval} presents the model's performance under various weather conditions, such as sunny, rainy, foggy, and stormy situations. Despite the imbalanced representation in the dataset, where sunny conditions dominate and extreme weather like sleet and snow are underrepresented, the model delivers reliable predictions. The MAE values remain relatively stable across all weather conditions, demonstrating the model's robustness even in less frequent and challenging scenarios. Notably, Tab.~\ref{tab:diff_weather_eval} also shows that even when certain weather conditions, such as storm and snow, are absent from the training dataset, the model still achieves good results on the corresponding test data. This highlights the strong generalization ability of the xTP-LLM model.

\subsubsection{zero-shot testing analysis}

\begin{table}[!h]
  \footnotesize
  \centering
  \begin{tabular}{c|ccc|ccc|ccc}
    \toprule 
\centering \multirow{2}{*}{Model} & \multicolumn{3}{c|}{CATraffic Zero-shot} & \multicolumn{3}{c|}{TaxiBJ (Inflow)} & \multicolumn{3}{c}{TaxiBJ (Outflow)} \\
   & RMSE  & MAE   &  MAPE(\%)  & RMSE  & MAE   &  MAPE(\%)  & RMSE  & MAE   &  MAPE(\%) \\
\midrule
    Llama2-7B-chat  & 273.02   & 226.64  & 126.00 &  255.00  &  151.57  &  172.60  &  218.44  &  135.39  & 151.96 \\ 
    Llama2-13B-chat & 234.73  &  187.37  & 146.08 & 230.55  &  129.39 &  184.90  &  234.14  &  144.24  &  143.47 \\
    Llama2-70B-chat & 222.02 & 168.04 & 122.72  &  205.67  &  123.46  &  136.83  &  209.05  &  125.39  & 139.89 \\    
    GPT-3.5-turbo  & 168.93  & 135.34  & 74.49 &  150.71  &  95.08  &  196.00  &  150.91  &  94.77  & 195.31 \\
    GPT-4  &  121.02 &  103.32 &  61.85 &  91.86  &  57.05  &  165.40  &  99.05  &  59.50  & 137.21 \\
    xTP-LLM  &\textbf{46.56}  &\textbf{29.73}  &\textbf{9.21} &  \textbf{54.82} &  \textbf{31.94}  &  \textbf{31.84}  &  \textbf{54.59}  &  \textbf{30.96}  & \textbf{31.20} \\
    \bottomrule
  \end{tabular}
  \caption{Zero-shot performance comparison of our proposed xTP-LLM with other LLMs without fine-tuning. 
We conducted tests on both the CATraffic zero-shot dataset and the TaxiBJ dataset to compare the zero-shot capabilities of different models. The results show that our model outperforms other LLMs significantly on both unseen datasets.}
  \label{tab:zeroshot_tab}
\end{table}

\begin{figure}[!h]
        \centering
        \includegraphics[width=\linewidth]{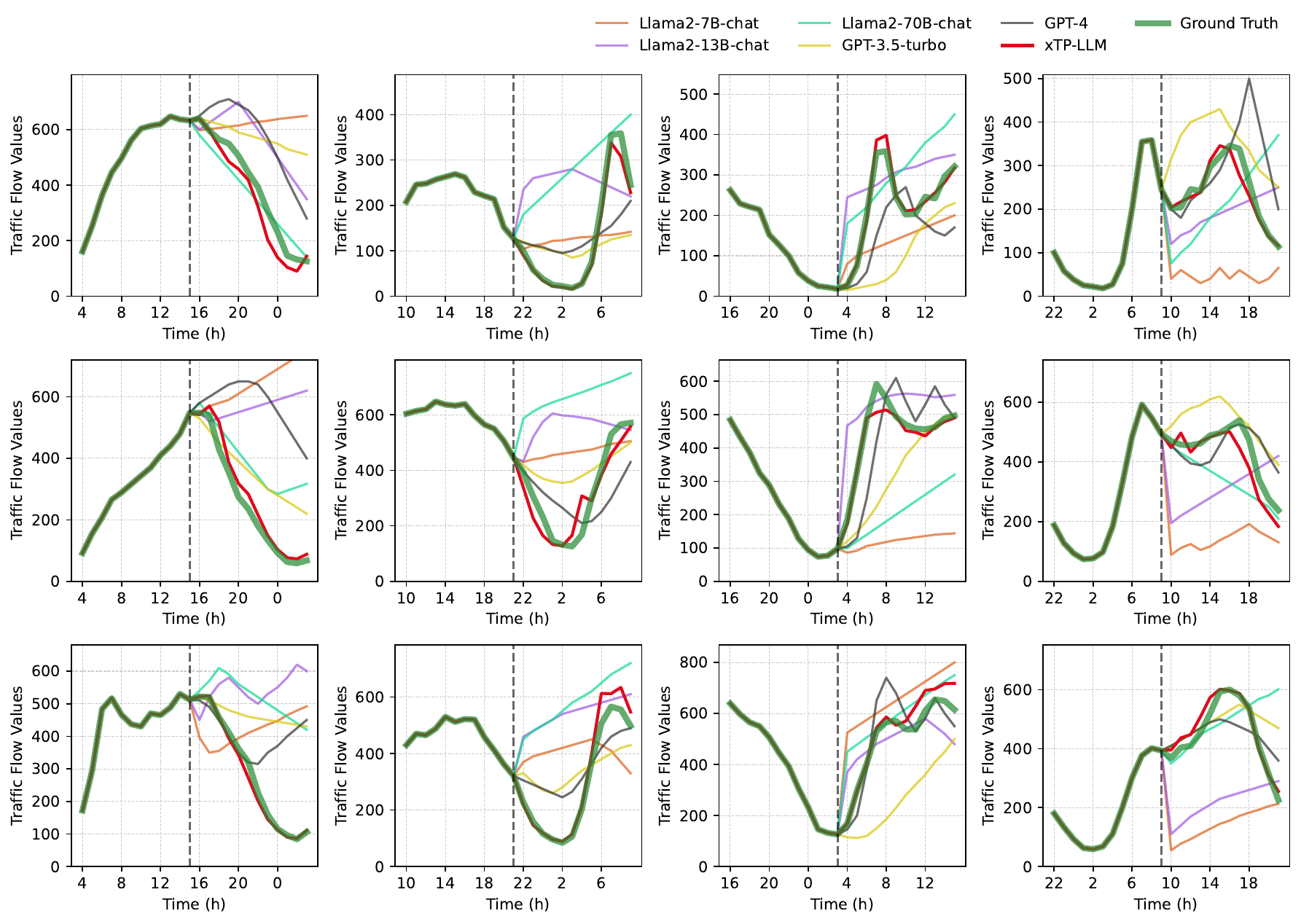}
        \caption{Visualization of traffic flow prediction results on CATraffic zero-shot dataset. We randomly select test samples of three sensors with four time periods and visualized the prediction results of our proposed xTP-LLM, Llama2-chat models with three different sizes, GPT-3.5-turbo and GPT-4. The results show that our fine-tuned model performs significantly better than other LLMs.}
\label{fig:zeroshot_vis}
\end{figure}

Large language models are well known for their excellent zero-shot capabilities.  In this section, we delve into the generalization of our proposed xTP-LLM and compare it with other large language models. We conducted zero-shot experiments using two datasets: the first is our proposed CATraffic-based zero-shot dataset, and the other is the taxiBJ dataset \cite{zhang2017deep}. The TaxiBJ dataset comprises taxicab GPS data and meteorology data in Beijing from four time intervals within 2013-2016, focusing on the inflow/outflow prediction task, which is different from the traffic volume prediction task in CATraffic. To ensure a fair comparison, we reorganized the taxiBJ dataset into the same format as our CATraffic dataset. The overall results are presented in Table~\ref{tab:zeroshot_tab}, which indicate that our proposed model exhibits superior performance across all three tasks compared to the original Llama2 series models, as well as GPT-3.5-turbo and GPT-4. For the CATraffic zero-shot dataset, our xTP-LLM model exhibits notable improvements over the best-performing comparative model (GPT-4) by 61.53\%, 71.22\%, and 85.11\% in RMSE, MAE, and MAPE, respectively. On the TaxiBJ dataset, our model achieves substantial performance enhancements in both inflow and outflow tasks. Compared to GPT-4, xTP-LLM shows improvements of 40.32\% in RMSE, 44.01\% in MAE and 80.75\% in MAPE in inflow prediction task, and improvements of 44.89\% in RMSE, 48.13\% in MAPE and 77.26\% inMAPE in outflow prediction task. These results demonstrate superior zero-shot learning capability of our xTP-LLM on the cross-domain datasets, even with a smaller scale of parameters (7 billion for xTP-LLM, compared to 70 billion for Llama2-70B-chat, 173 billion for GPT-3.5, and approximately 1.76 trillion for GPT-4). While large language models like GPT-4 and Llama2-70B-chat exhibit strong generalization capabilities in general domains such as conversation and translation, they generally do not perform well in the traffic prediction domain when applied in a zero-shot learning setting.

We also provide visualization results in Figure~\ref{fig:zeroshot_vis}, showing 12 test samples from three different sensors in four different time periods. The results highlight that our model not only captures the traffic trends of new scenarios effectively but also delivers accurate prediction values. In contrast, Llama2 series of LLMs struggle to capture the dynamic pattern of traffic flow; while GPT-3.5-turbo and GPT-4 demonstrate the ability to describe some general trends in traffic, they perform inadequately in capturing nuances of variation, hindering their ability to provide accurate predictions. We attribute these discrepancies to our systematic prompt design and effective model fine-tuning. Prompt inputs effectively describe the task context and align information from different modalities into a unified representation, while fine-tuning infuses crucial expert knowledge into the model, thereby substantially enhancing its performance in traffic prediction. Our experiments further validate that instruction fine-tuning allows smaller models to excel in specific domains, often outperforming larger models.

\subsection{Case Studies for Explainable Predictions}
\subsubsection{Result-explainable studies}
\begin{table}[!h]
  \begin{tabularx}{\textwidth}{cc}
    \toprule
    \textbf{Prediction} & \textbf{Explanation} \\
    \midrule
    \begin{minipage}[h]{0.4\columnwidth}
		\centering
		\includegraphics[width=0.85\linewidth]{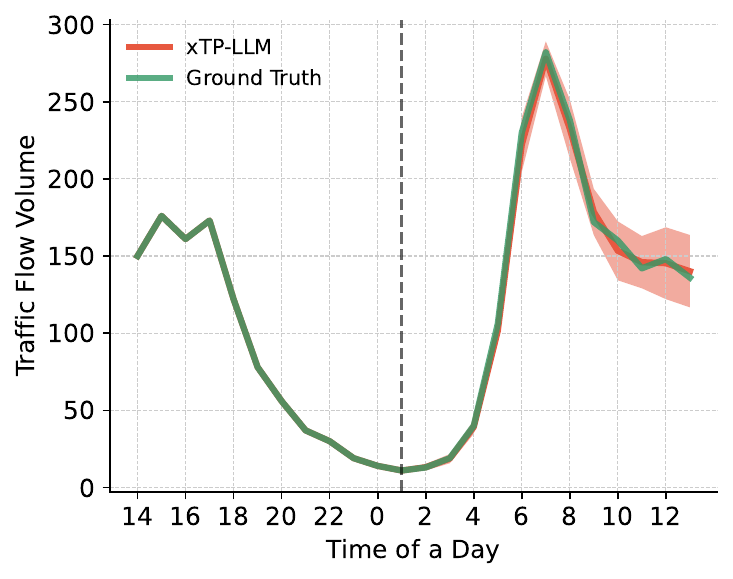}
	\end{minipage}
    & \begin{minipage}[h]{0.55\columnwidth}
         \textbf{(a):} Traffic volume is expected to be low before dawn, then it is expected to increase gradually throughout the day. Since today is Wednesday, weekday patterns suggest \sethlcolor{yellow!50}\hl{higher traffic during rush hours and lower traffic during non-peak times.} Nearby transportation areas and commercial districts will contribute to an increase in traffic volume during certain periods. Sunny weather can also contribute to increased traffic.
    \end{minipage}
    \\
    \midrule
    \begin{minipage}[h]{0.4\columnwidth}
		\centering
		\includegraphics[width=0.85\linewidth]{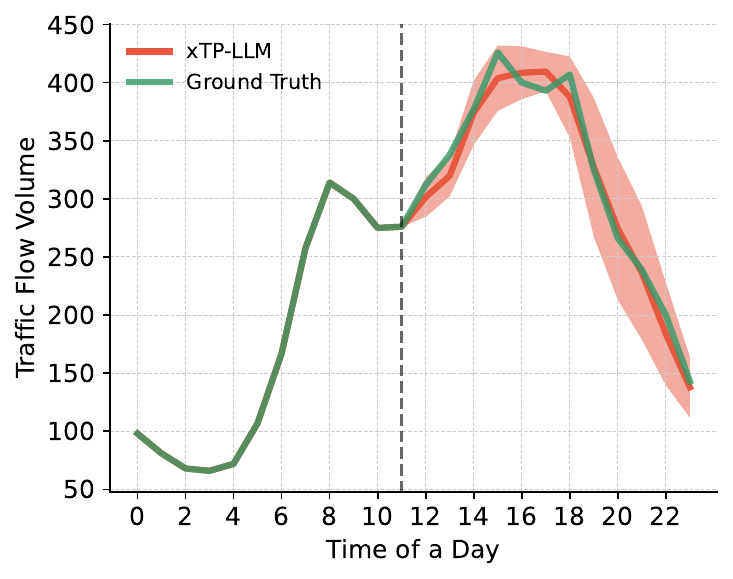}
    \end{minipage}
    & 
    \begin{minipage}[h]{0.55\columnwidth}
       \textbf{(b):} Traffic volume is expected to \sethlcolor{yellow!50}\hl{increase and then decrease} in the next 12 hours. Sunny weather with a comfortable temperature will attract more day travelers. The onset of Sunday afternoon rush hour will further contribute to increased traffic. With no known holidays or events, historical data suggest that traffic volume will resemble previous Sundays, with a slight increase during the afternoon rush hour followed by a gradual decline at night.
    \end{minipage} 
    \\
    \midrule
    \begin{minipage}[h]{0.4\columnwidth}
		\centering
		\includegraphics[width=0.85\linewidth]{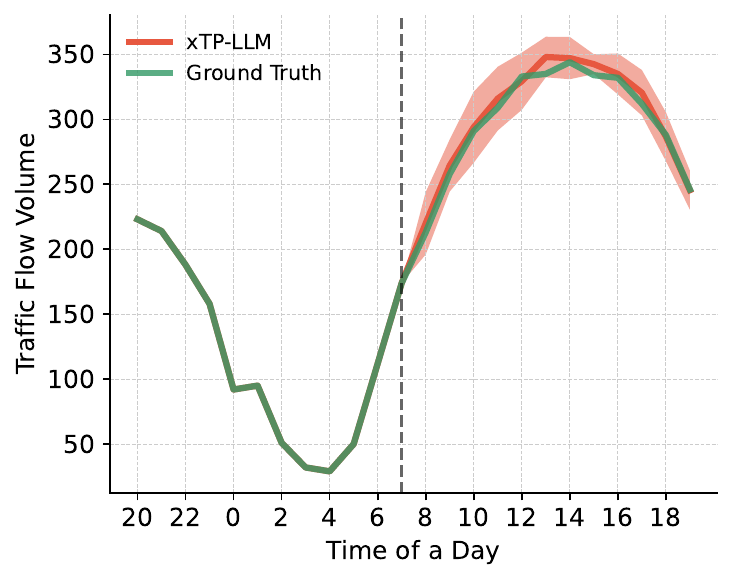}
	\end{minipage}
    & \begin{minipage}[h]{0.55\columnwidth}
        \textbf{(c):} Traffic volume is expected to \sethlcolor{yellow!50}\hl{steady increase in the rainy morning and then decrease in the afternoon}. Activities at the morning rush hour in residential and commercial areas are likely to contribute to the initial rise. The rain might cause congestion, leading to a peak in volumes around midday. As the day progresses, the volume is expected to gradually decrease as people return home.
    \end{minipage}
    \\
    \midrule
    \begin{minipage}[h]{0.4\columnwidth}
		\centering
		\includegraphics[width=0.85\linewidth]{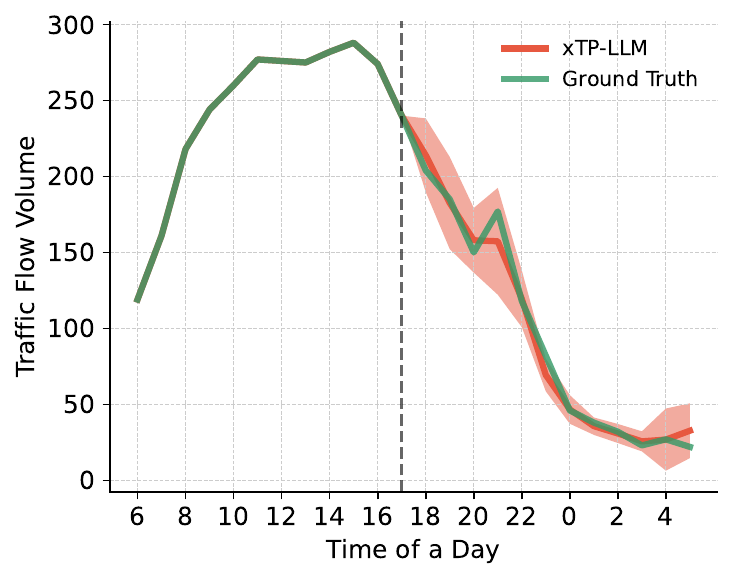}
    \end{minipage}
    & 
    \begin{minipage}[h]{0.55\columnwidth}
        \textbf{(d):} Traffic volume is expected to \sethlcolor{yellow!50}\hl{decrease significantly} over the next 12 hours considering a rainy Christmas Day. As it is a holiday, commuter and commercial activities are minimal, leading to reduced traffic. Additionally, the rainy weather may further deter people from traveling or driving, contributing to the decline in traffic volumes. The focus on indoor celebrations and festivities is likely to keep the roads quiet at night.
    \end{minipage}
    \\
    \bottomrule
  \end{tabularx}
  \caption{Examples of the explanations. We show the ground truth and our predicted result in the figure, with the 95\% confidence interval represented in light red; and the generated explanations on the right, with predicted traffic trends highlighted in yellow. Since the original outputs are too long, we use ChatGPT to summarize them. A complete example can be found in Appendix~\ref{sec:explanatory_case}.}
   \label{tab:explanation}
\end{table}

To illustrate the reliability and accountability of our model's predictions, we report four cases with different prediction time slots and external factors in Table~\ref{tab:explanation}. In each example, the ground truths and predictions are displayed in the figure on the left (light red areas indicate the 95\% confidence interval), with the corresponding explanatory texts on the right. The original result explanations are too long, so we used chatGPT to summarise them into a brief paragraph for presentation purposes. A complete output can be found in Appendix~\ref{sec:explanatory_case} (corresponding to example (d) in Table~\ref{tab:explanation}). 

Our proposed xTP-LLM consistently delivers accurate traffic flow predictions across various time periods, including weekdays, weekends, and holidays like Christmas. Notably, our model not only provides precise forecasts but also offers insightful explanations, considering a wide range of external factors. For instance, in example (a), our model predicts an initial increase followed by a decrease in traffic flow, aligning with typical weekday patterns. Moreover, in example (c), our model identifies rainy weather as an impact factor for traffic congestion, leading to a delayed peak traffic flow. Furthermore, example (d) highlights our model's ability to factor in holiday impacts on traffic flow. 

While our xTP-LLM excels in leveraging diverse external data for robust predictions, it faces challenges in capturing nuanced flow fluctuations, such as those occurring between 14-18 hours in Example 2 and 20-22 hours in Example (d). These complexities may stem from the dynamic nature of human activities and the intricacies of transportation systems. Nevertheless, although not good at capturing the full details of the traffic flow changes, our xTP-LLM provides textual rationales for its forecasts, which shows a high transparency and interpretability.

\subsubsection{What-if Analysis}
To investigate the generalization and conditional reasoning abilities of the framework, this section examines how prediction results change when influenced by external factors, such as traffic accidents and inclement weather. For the situational analysis, we consider two hypothetical scenarios: Case 1 involves a nearby traffic accident occurring at 10 p.m., and Case 2 involves a severe sandstorm breaking out at 9 a.m. These conditional descriptions are incorporated into the input prompt, while the other sections of the prompt remain consistent with the original prediction tasks. We then compare the normal predictions, what-if predictions, and the ground truth, as shown in Fig \ref{tab:wi}.

\begin{table}[!h]
  \begin{tabularx}{\textwidth}{cc}
    \toprule
    \textbf{Prediction} & \textbf{Explanation} \\
    \midrule
    \begin{minipage}[h]{0.5\columnwidth}
		\centering
		\includegraphics[width=0.85\linewidth]{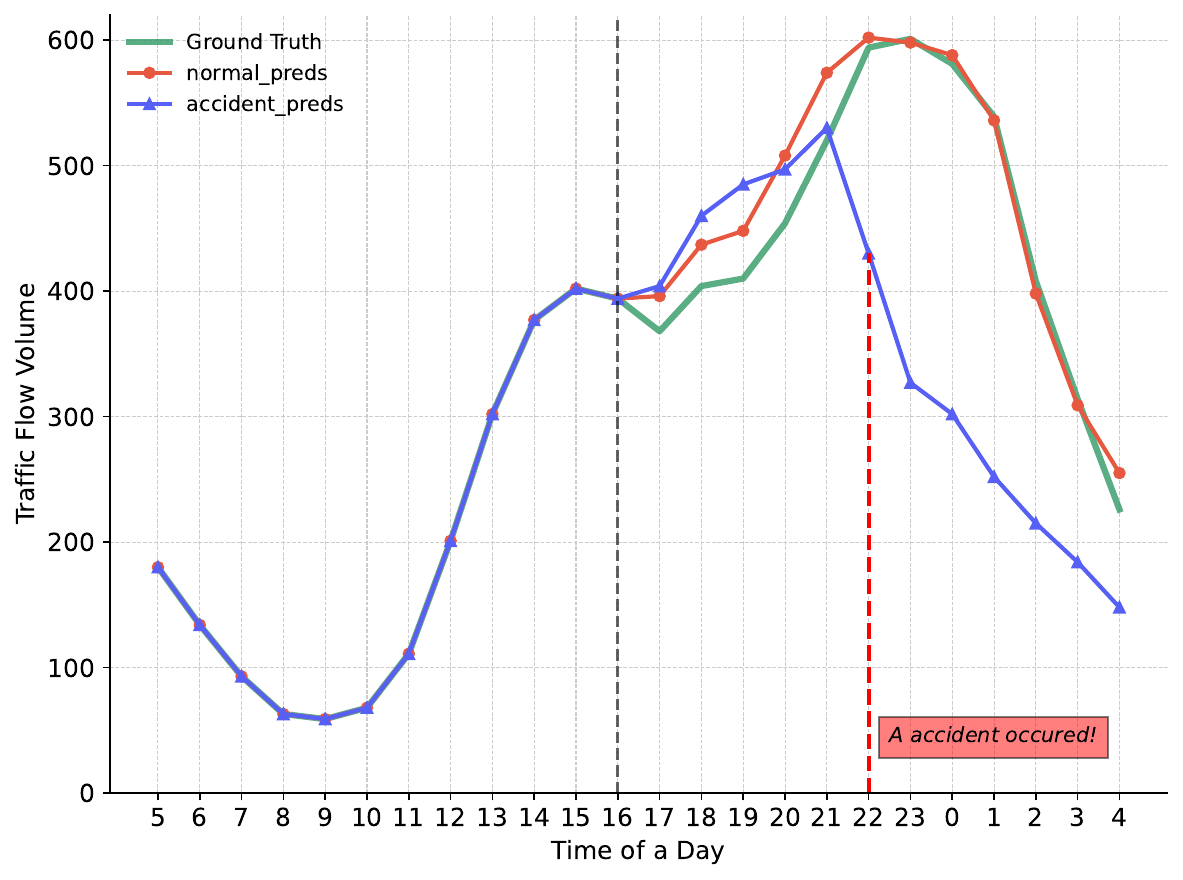}
	\end{minipage}
    & 
    \begin{minipage}[h]{0.45\columnwidth}
         \textbf{Sudden Accident:}  ...Important! A serious traffic accident occurred on this road at 10 PM!...
         
         \textbf{Explanation:} ...The traffic volume in the next 12 hours will increase due to the post-accident recovery. The accident had a significant impact on traffic flow \sethlcolor{yellow!50}\hl{causing delays and changes in traffic patterns}. After the accident, the number of cars \sethlcolor{yellow!50}\hl{decreased} but still remained at a relatively high level.
    \end{minipage}
    \\
    \midrule
    \begin{minipage}[h]{0.5\columnwidth}
		\centering
		\includegraphics[width=0.85\linewidth]{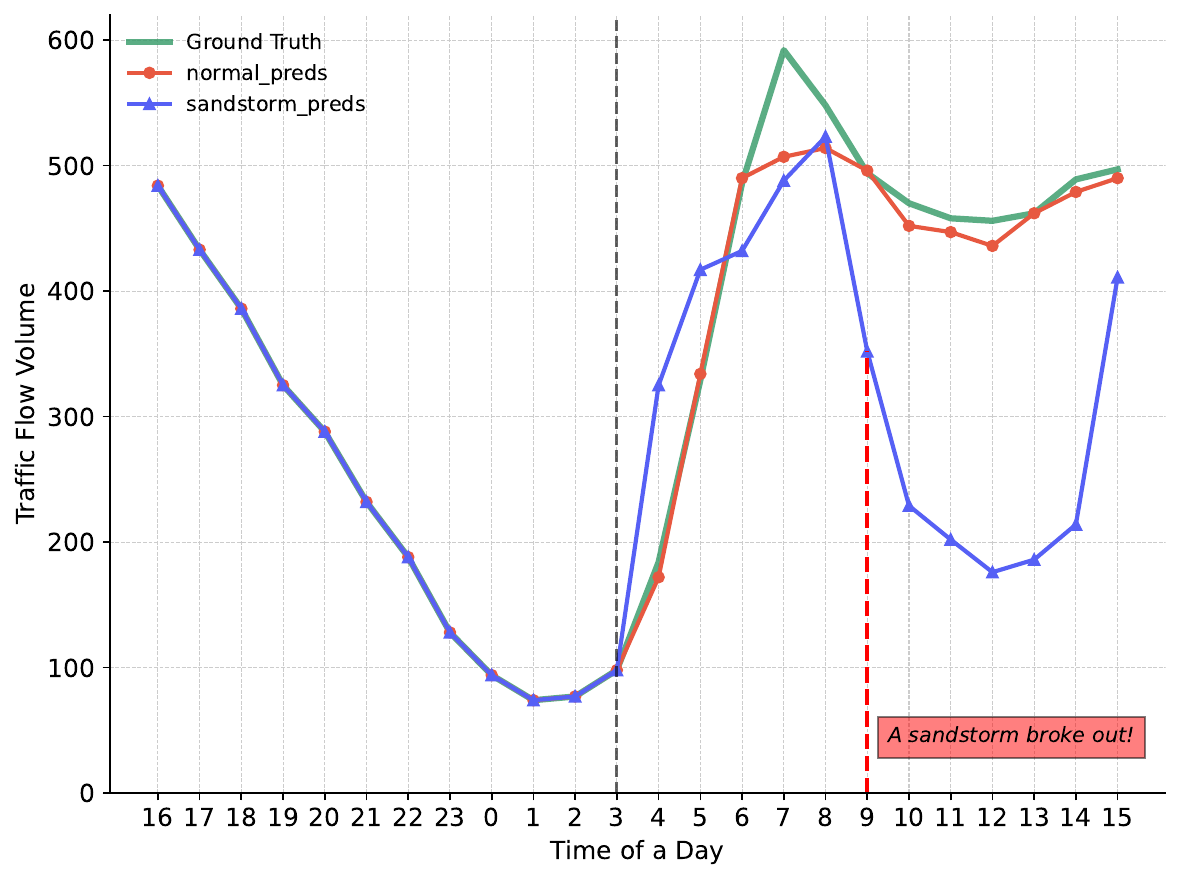}
    \end{minipage}
    & 
    \begin{minipage}[h]{0.45\columnwidth}
       \textbf{Sudden Weather Event:}  ...Important! A severe sandstorm broke out at 9 AM!...
       
       \textbf{Explanation:} ...The sandstorm causes \sethlcolor{yellow!50}\hl{a sharp drop in traffic after 9 AM}, with volumes significantly lower than normal due to hazardous conditions, recovering slightly in the afternoon as the storm eases...
    \end{minipage} 
    \\
    \bottomrule
  \end{tabularx}
  \caption{Examples of what-if analysis.}
   \label{tab:wi}
\end{table}

The traffic system is highly dynamic and complex, interacting with human factors and real-time external conditions. From the experimental results, we observe that when a traffic accident occurs at a specific moment, the nearby traffic volume decreases due to the congestion caused by the accident, dropping from 530 to 430. The model can interact with real-time incidents, avoiding prediction delays and incorrect analysis. Similarly, when severe weather events like a sandstorm occur, the reduced visibility and road passability lead to a sharp decline in the number of vehicles passing through the area, from 532 to 352. However, over time, traffic flow gradually recovers to normal levels in the region. This indicates that our model provides timely feedback to sudden weather changes, and its inherent reasoning and analytical capabilities help make more reasonable predictions.

\section{Conclusion and Future Work}
In conclusion, our research introduces xTP-LLM, a novel traffic prediction model designed for both accuracy and interpretability. By incorporating multi-modal inputs and employing language-based representations, xTP-LLM achieves competitive performance compared to state-of-the-art models while offering insightful explanations into its predictions. xTP-LLM's language-based framework, coupled with spatial-temporal alignment instructions, provides a transparent and adaptable approach suitable for various urban prediction tasks. In general, our work contributes to the advancement of effective and reliable traffic prediction methods, essential for informed decision-making in urban transportation planning and management. 

In the future, we aim to delve into methods that enable LLMs to harness spatial information more effectively and to grasp how different sensors are related spatially. This will help models to make better predictions by considering data from nearby sensors. Moreover, further external factors can be considered, such as traffic accidents, human activities, and big events, contributing to more accurate predictions. Additionally, exploring the development of LLM systems tailored for urban brains is also a very interesting but challenging topic. This entails integrating city-level data into LLMs to tackle various downstream tasks like urban planning, traffic management, and pollution control, etc. Achieving these involves the challenge of enabling LLMs to efficiently utilize city-level multi-modal data, alongside the need for substantial computational resources and exceptional engineering capabilities. 

\section*{Data and Code Availability}

The training and evaluation code can be accessed at GitHub (\href{https://github.com/Guoxs/xTP-LLM}{https://github.com/Guoxs/xTP-LLM}). A portion of the CATraffic dataset, covering traffic flow data for 100 traffic nodes over three months, is available at Google Cloud, and the download link can be found in our GitHub repository. The complete dataset will be made available later through the same link after further organization.

\section{Acknowledgements}
This research is funded by multiple sources, including the National Natural Science Foundation of China under Grant 52302379, the Guangzhou Basic and Applied Basic Research Projects under Grants 2023A03J0106 and 2024A04J4290, the Guangdong Province General Universities Youth Innovative Talents Project under Grant 2023KQNCX100, and the Guangzhou Municipal Science and Technology Project 2023A03J0011.

\newpage

\bibliographystyle{R2T-LLM_revised_new}
\bibliography{R2T-LLM_revised_new}

\newpage

\appendix
\section{Appendix}
\subsection{Prompt Design}
\label{sec:prompt_design}

The prompt for xTP-LLM in traffic flow prediction is carefully designed, as shown in Table~4. The complete prompt contains both a system prompt and a user input prompt. The system prompt sets the role of the LLM, and it also contains a context knowledge part that provides additional background information in traffic flow prediction, as well as the CoT prompt that guides LLMs through the human reasoning process. The system prompt remains the same throughout the dataset, and the change parts are the user input prompt as well as the corresponding ground truth.

\begin{table}[!h]
\footnotesize
\begin{tabular}{p{\textwidth}}
\toprule
\textbf{System Prompt:} \\
You are an expert traffic volume prediction model, that can predict the future volume values according to spatial temporal information. We want you to perform the traffic volume prediction task, considering the nearby environment and historical traffic volume data.\\

\textbf{Context knowledge:} \\
Context knowledge you could consider:
\begin{itemize}
    \item Traffic volume: the number of vehicles passing a specific region in an hour.
    \item Traffic pattern characteristic: Traffic flow patterns in a city are influenced by various area attributes. Also, traffic volume has a periodic daily and weekly pattern.
    \item Spatial-temporal factors correlation: Traffic flow in an area will be affected by its nearby infrastructures, during specific periods for different areas. You should think about how the volume will change in a specific area, during a specific time. For example, 
        \begin{itemize}
            \item  Airports, and train stations - increased volume on weekends and holidays.
            \item  Residential areas - more activities during morning and evening rush hours.
            \item  Commercial areas - busy during lunch hours and after-work periods.
            \item  Educational locations - high volume during peak hours near schools.
        \end{itemize}
\end{itemize}

\textbf{Chain of Thought: }\\
Think carefully about the following questions about how spatial-temporal factors affect traffic flow.
\begin{itemize}
    \item What is the attribute of this area and what is the predicted time zone located in special periods (like rush hours, weekdays, weekends, and holidays)?
    \item What are the traffic patterns of this area, and what is the change in different time slots?
    \item What is the historical temporal trend according to temporal information, considering the weekdays,  around holidays?
\end{itemize} 
\\
\midrule
\textbf{Input Prompt:} \\
Some important information is listed as follows:
\begin{itemize}
    \item Location: District 3 in Yolo, California, USA, along the US50-E freeway, lane 4, direction of eastbound.
    \item Today's weather: Sunny. The temperature is 6.0$^{\circ}$C and the visibility is 10.0 miles.
    \item Region information: including transportation areas, commercial areas, and educational areas within a range of 5 km.
    \item Current time: 3 PM, 2018-2-19, Monday, Washington's Birthday.
    \item Traffic volume data in the past 12 hours were 19, 44, 98, 150, 156, 178, 208, 246, 248, 257, 263 and 269, respectively.
\end{itemize}

According to the above information and careful reasoning, please predict traffic volumes in the next 12 hours (from 4 PM to 3 AM). Format the final answer in a single line as a JSON dictionary like: {Traffic volume data in the next 12 hours: [V1, V2, V3, V4, V5, V6, V7, V8, V9, V10, V11, V12]}.\\
\midrule
\textbf{Ground Truth: }\\
{Traffic volume data in the next 12 hours: [262, 229, 221, 214, 152, 127, 100, 58, 38, 25, 22, 18]}. \\
\bottomrule
\caption{Details of our designed prompt in xTP-LLM for traffic flow prediction task, including system prompt, input prompt, as well as the ground truth.}
\end{tabular}
\label{tab:prompt_details}
\end{table}

\subsection{Explanatory Example}
\label{sec:explanatory_case}

A complete example of interpretative output is given in Table~5. This example shows the traffic flow prediction in the next 12 hours at 5 PM on Christmas Day 2019. By adding explanatory demand instructions to the original prompt (bold text in \textit{Input Prompt} of Table~5), xTP-LLM can output both the prediction results as well as the explanatory text. It can be found that our model is capable of adjusting the traffic flow prediction strategy by taking into account holidays and the impact of weather on human activities, which shows that our model is highly reliable and accountable.

\begin{table}[!h]
\footnotesize
\begin{tabular}{p{\textwidth}}
\toprule
\textbf{System Prompt:} (The same as in Table~4) ... \\
\midrule
\textbf{Input Prompt:} \\
Some important information is listed as follows:
\begin{itemize}
    \item District 3 in Yolo, California, USA, along the US50-W freeway, lane 3, direction of westbound. 
    \item Today's weather:  Rain. The temperature is 8.1$^{\circ}$C and the visibility is 9.7 miles. 
    \item Region information: including transportation areas, commercial areas and educational areas within a range of 5 km.
    \item Current time: 5 PM, 2019-12-25, Wednesday, Christmas Day.
    \item Traffic volume data in the past 12 hours were 119, 161, 218, 244, 260, 277, 276, 275, 282, 288, 274 and 240, respectively.
\end{itemize}

According to the above information and careful reasoning, please predict traffic volumes in the next 12 hours (from 2 AM to 1 PM) \textbf{and explain it}. Format the final answer in a single line as a JSON dictionary like: \{Traffic volume data in the next 12 hours: [V1, V2, V3, V4, V5, V6, V7, V8, V9, V10, V11, V12],\textbf{ Explanation: xxx}\}. \textbf{Please think step by step.}\\
\midrule
\textbf{xTP-LLM's Response:}\\
\{"Traffic volume data in the next 12 hours": [214, 183, 158, 157, 119, 69, 47, 36, 31, 26, 27, 33], \\
    "Explanation": I will provide a step-by-step explanation for predicting traffic volumes in the next 12 hours based on the given information.\\
    1. Time and Date: It is 5 PM on a Wednesday, December 25th, 2019, which is Christmas Day, with rainy weather and good visibility.\\
    2. Historical Traffic Volume: In the past 12 hours, the traffic volumes were: [119, 161, 218, 244, 260, 277, 276, 275, 282, 288, 274, 240].\\
    3. Current Situation: Given the holiday, Christmas Day, and the rainy weather, traffic patterns may differ from regular weekdays.\\
    4. Christmas Day Factor: On holidays, traffic volume might be lower due to decreased commuter and commercial activities. People may stay indoors for celebrations.\\
    5. Rain Impact: The rain could further reduce traffic volume as people may prefer to avoid driving in adverse weather conditions.\\
    6. Traffic Trend Prediction: \\
    \quad\quad    - The evening hours (6 PM to 9 PM): Traffic volume is likely to be significantly lower than usual as people prioritize holiday celebrations. \\
    \quad\quad    - Late-night hours (10 PM to 5 AM): Traffic volume may remain low throughout the night.
\} \\
\midrule
\textbf{ChatGPT Summary:} \\
Considering the current time at 5 PM on a rainy Christmas Day, traffic volume is expected to decrease significantly over the next 12 hours. As it is a holiday, commuter and commercial activities are minimal, leading to reduced traffic. Additionally, the rainy weather may further deter people from traveling or driving, contributing to the decline in traffic volumes. The focus on indoor celebrations and festivities is likely to keep the roads quiet throughout the night.
\\
\bottomrule
\vspace{6pt}
\caption{An example of the complete response with explanatory output of our proposed xTP-LLM. This example corresponds to the example (d) in Table~\ref{tab:explanation}.}
\end{tabular}
\label{tab:explanatory_example}
\end{table}

\end{document}